\documentclass[10pt, a4paper, onecolumn, twoside]{amsart}

\usepackage{titlesec, color, amsthm}
\definecolor{blue1}{rgb}{0,0,0}
\renewcommand\thesection{\arabic{section}}

\titleformat{\section}[hang]{\color{blue1}\large\bfseries\sffamily}{\thesection}{0mm}{. }[]
\titleformat{\subsection}[hang] {\color{blue1}\bfseries\sffamily}{\thesubsection}{0em}{. }[]
\titleformat{\subsubsection}[hang] {\color{blue1}\sffamily}{\thesubsubsection}{0em}{. }[]
\titlespacing*{\section}{1em}{3.5ex plus .2ex minus .2ex}{1ex plus .2ex}
\titlespacing*{\subsection}{0em}{3ex plus .2ex minus .2ex}{1ex plus .2ex}
\titlespacing*{\subsubsection}{0em}{3ex plus .2ex minus .2ex}{1ex plus .2ex}

\renewenvironment{abstract}{{\color{blue1}\small\bfseries Abstract.}\footnotesize}{\par \vskip .1in}
\makeatletter
\def\@setauthors{
\begingroup 
\def \thanks{\protect\thanks@warning}
\trivlist \centering\footnotesize \@topsep30\p@\relax \advance\@topsep by -\baselineskip
\item\relax \author@andify \authors \def\\{\protect\linebreak} {\color{blue1}\large\authors} \endtrivlist \endgroup}
\def\@settitle{\centering{\color{blue1} \Large \bfseries \bfseries \@title \par}}
\makeatother
\usepackage[normal, small, up, textfont=it]{caption}
\usepackage{url}
\usepackage{graphicx}
\usepackage{subfigure} 
\usepackage{algorithm}
\usepackage{algorithmic}
\usepackage{bm}
\usepackage{amsmath}
\usepackage{amssymb}
\usepackage{enumerate}
\usepackage{mathtools}

\newcommand{\longtitle}{{Unifying local and non-local signal processing with graph CNNs}}

\newcommand{\GPlong}{{Gilles~Puy}} 
\newcommand{\SKlong}{{Sr\dj{}an~Kiti\'c}} 
\newcommand{\PPlong}{{Patrick~P\'erez}} 
\newcommand{\Technicolor}{{Technicolor, 975 Avenue des Champs Blancs, 35576 Cesson-S\'evign\'e, France}}

\title[\longtitle]{\longtitle}
\author{\GPlong}
\author{\SKlong}
\author{\PPlong}
\thanks{\GPlong, \SKlong\ and \PPlong\ are with \Technicolor.}

\newcommand{\PP}[1]{{\color{blue}[\textbf{PP}: #1]}}
\newcommand{\GP}[1]{{\color{blue}[\textbf{GP}: #1]}}
\newcommand{\add}[1]{{\color{black} #1}}

\newcommand{\Graph}{\ensuremath{\set{G}}}

\newcommand{\Lap}{\ensuremath{\ma{L}}}
\newcommand{\Fou}{\ensuremath{\ma{U}}}
\newcommand{\Eig}{\ensuremath{\ma{\Lambda}}}

\newcommand{\eig}{\ensuremath{\lambda}}

\newcommand{\Sig}{\ensuremath{\ma{X}}}
\newcommand{\sigVec}{\ensuremath{\vec{x}}}

\newcommand{\bias}{\ensuremath{b}}

\newcommand{\nbVert}{\ensuremath{n}}

\newcommand{\nbFeat}{\ensuremath{m}}
\newcommand{\gdeg}{\ensuremath{d}}

\newcommand{\Rbb}{\ensuremath{\mathbb{R}}}

\renewcommand{\leq}{\ensuremath{\leqslant}}
\renewcommand{\geq}{\ensuremath{\geqslant}}

\DeclarePairedDelimiter\floor{\lfloor}{\rfloor}

\DeclareMathOperator*{\argmin}{argmin}

\newcommand{\scp}[2]{\ensuremath{\left\langle #1, #2 \right\rangle}}
\newcommand{\adjoint}{\ensuremath{{\intercal}}}

\newcommand{\norm}[1]{\ensuremath{\left\| #1\right\|}}
\newcommand{\abs}[1]{\ensuremath{\left| #1 \right|}}

\newcommand{\ma}[1]{\ensuremath{\mathsf{#1}}}
\renewcommand{\vec}[1]{\ensuremath{\bm{#1}}}
\newcommand{\diag}{\ensuremath{{\rm diag}}}

\newcommand{\set}[1]{\ensuremath{\mathcal{#1}}}

\newcommand{\Lab}{\textit{Lab}}
\newcommand{\ab}{\textit{ab}}
\newcommand{\ie}{\textit{i.e.}}
\newcommand{\eg}{\textit{e.g.}}
\renewcommand{\th}{\ensuremath{\text{th}}}

\newcommand{\suppl}{{the Appendix}}
\graphicspath{{figs/}}
\DeclareGraphicsExtensions{.pdf,.jpeg,.jpg,.eps,.png}

\begin{document} 

\maketitle
\begin{abstract} 
This paper deals with the unification of local and non-local signal processing on graphs within a single convolutional neural network (CNN) framework. Building upon recent works on graph CNNs, we propose to use convolutional layers that take as inputs two variables, a signal and a graph, allowing the network to adapt to changes in the graph structure. \add{In this article, we explain how this framework allows us to design a novel method to perform style transfer.}
\end{abstract}

\section{Introduction}

Convolutional neural networks (CNNs) have achieved unprecedented performance in a wide variety of applications, in particular for image analysis, enhancement and editing -- \eg, classification \cite{krizhevsky12}, super-resolution \cite{dong16}, and colorisation \cite{zhang16}. Yet standard CNNs can only handle signals that live on a regular grid, and each layer of a CNN only performs a local processing. Locality has already been identified as a limitation for classical signal processing tasks where powerful non-local methods have been proposed, such as patch-based methods for inpainting \cite{criminisi04} or denoising \cite{buades11,talmon2011transient}.
Regular CNNs do not allow such a non-local processing. Furthermore, the growing amount of signals collected on irregular grids, such as social, transportation or biological networks, requires extending signal processing from regular to irregular graphs \cite{shuman_SPMAG2013}.

Any CNN consists of a composition of convolutional and pooling layers. One should thus redefine both convolution and pooling to handle ``graph signals''. In this work, we use convolutional layers only and hence just concentrate on the generalisation of the convolution. One major challenge in this generalisation is to take into account the possible changes of the graph structure from one signal instance to another: nodes and edges can appear, disappear, and the edge weights can vary. 
For instance, the connections between the users (graph's vertices) in a social network change over time. 
It would be cumbersome to retrain a CNN each time a connection changes. In non-local signal processing methods, the situation is even more extreme as the graph is a construct whose edges typically capture similarities between different parts of the signal itself. In this case, the CNN must not be just robust to few variations in the graph structure but fully adapt to these variations. We propose here a solution to this challenge but passing two variables to the CNN: the signal itself, as usual, and the graph structure.

\textbf{Contributions} -- \add{We propose a graph CNN framework that takes as inputs two variables: a signal and a graph structure. This permits the adaptation of the CNN to changes in the structure of the graph on which the signal lives, even in the extreme case where this structure changes with the input signal itself. We also propose a unique way of defining convolutions on arbitrary graphs, in particular non-local convolutions, with application to a wide range of many different signal processing applications. Due to space constraint, we only present the use of graph CNNs for image style transfer in this article. We use a local CNN to capture and transfer local style properties of the painting to the photograph. We also use a non-local graph CNN to capture and transfer global style properties of the painting, as well as to preserve the content of the photograph. In addition, we show that this task can be done using only \emph{two random shallow} networks, instead of a trained regular deep CNN \cite{gatys16}. 

Let us mention that additional experiments in \suppl\ demonstrate the effectiveness and versatility of our framework on other kinds of signals (greyscale images, color palettes, and speech signals) and tasks (color transfer and denoising). In particular, the experiments show that it is possible to identify the optimal mixing of local and non-local signal processing techniques by learning.}

\section{Graph CNN}

\subsection{State-of-the-art methods}

In this section, we review different existing solutions to generalise CNNs to signal living on graphs. The reader can refer to \cite{bronstein16} for a detailed overview. We restrict our attention here to the solutions the most closely connected to ours. 

A first approach to redefine convolution for graph signals is to work in the spectral domain, for which we need to define the graph Fourier transform. To introduce this transform, we consider an undirected weighted graph\footnote{A graph $\Graph$ is a set of $n$ vertices, a set of edges $\set{E}$ and a weighted adjacency matrix $\ma{W}=[\ma{W}_{ij}]\in\mathbb{R}_+^{n\times n}$, with $\ma{W}_{ij}>0$ iff $(i,j) \in \set{E}$. In this paper, we consider directed graphs unless explicitly stated. The matrix $\ma{W}$ is thus not symmetric in general.}
$\Graph$ with graph Laplacian denoted by $\Lap \in \Rbb^{\nbVert \times \nbVert}$. For example, $\Lap$ can be the combinatorial graph Laplacian $\Lap = \ma{D} - \ma{W}$, or the normalised one $\Lap = \ma{I} - \ma{D}^{-1/2} \ma{W} \ma{D}^{-1/2}$, where $\ma{I}$ is the identity matrix and $\ma{D} \in \Rbb^{\nbVert \times \nbVert}$ is the diagonal degree matrix with entries
$d_i = \sum_{j=1}^{\nbVert}\ma{W}_{ij}$ ~\cite{chung_book1997}. 
The matrix $\Lap$ is real symmetric and positive semi-definite. Thus, there exists a set of orthonormal eigenvectors $\Fou \in \Rbb^{\nbVert \times \nbVert}$ and real eigenvalues $0 = \eig_1 \leq \ldots \leq \eig_n$ such that $\Lap = \Fou \Eig \Fou^\adjoint$, where $\Eig = \diag(\eig_1, \ldots, \eig_n) \in \Rbb^{\nbVert \times \nbVert}$. The matrix $\Fou$ is viewed as the graph Fourier basis~\cite{shuman_SPMAG2013}. 

For any signal $\sigVec \in \Rbb^{\nbVert}$ defined on the vertices of $\Graph$, $ \hat{\sigVec} = \Fou^\adjoint \sigVec$ is its graph Fourier transform. One way to define convolution on $\Graph$ with a filter $\vec{h} \in \Rbb^\nbVert$ is by filtering in the graph Fourier domain:
\begin{align}
\sigVec \star \vec{h} 
= \Fou \; (\hat{\vec{h}} \odot \hat{\sigVec})
= \Fou \ma{H} \Fou^\adjoint \sigVec,
\end{align}
where $\odot$ denotes the entry-wise multiplication and $\ma{H} = \diag(\hat{\vec{h}}) \in \Rbb^{\nbVert \times \nbVert}$. In the context of graph CNN, it is the approach chosen in \cite{bruna14}. This approach however has several drawbacks: Computing $\Fou$ is often intractable for real-size graphs; Matrix-vector multiplication with $\Fou$ is usually slow (there is no fast graph Fourier transform); This definition does not allow  variations in the graph structure as the matrix $\Fou$ is impacted by any such change; The number of filter coefficients to learn is as large as the size of the input signal. The subsequent work \cite{henaff15} solves this last issue by imposing that the filter lives in the span of a kernel matrix $\ma{K}^{\nbVert \times \tilde{\nbVert}}$ with $\tilde{\nbVert} \leq \nbVert$.

To overcome the computational issues of the spectral approach, a known trick in the field of graph signal processing is to define a filter as a polynomial of the graph eigenvalues \cite{hammond11}. Let $\hat{h} \colon \Rbb \rightarrow \Rbb$ be a polynomial of degree $m\geq 0$: $\hat{h}(t) = \sum_{i=0}^m \alpha_i t^i$, with $\alpha_0, \ldots, \alpha_m \in \Rbb$ and consider the filter $\hat{\vec{h}} = (\hat{h}(\eig_1), \ldots, \hat{h}(\eig_\nbVert))^\adjoint$. One can easily prove that spectral filtering with $\hat{\vec{h}}$ satisfies
\begin{align}
\label{eq:conv_poly}
\sigVec \star \vec{h} = \Fou \; (\hat{\vec{h}} \odot \hat{\sigVec}) = \sum_{i=0}^m \alpha_i \Lap^i \sigVec.
\end{align}
This expression involves only computations in the vertex domain through matrix-vector multiplications with $\Lap$. As the Laplacian is usually a sparse matrix, filtering a signal with a polynomial filter is fast. This is the approach adopted by \cite{defferrard16} and \cite{kipf16} in their construction of graph CNNs. Beyond the computational improvements, the number of coefficients to learn is also reduced: $m$ instead of $\nbVert$. Furthermore, the localisation of the filter in the vertex domain is exactly controlled by the degree of the polynomial \cite{hammond11, defferrard16}. Yet these polynomial filters are not entirely satisfying. Indeed, for, \eg, a graph modelling a regular lattice, polynomial filters are isotropic unlike those in regular CNNs for images -- where the underlying graph is a regular lattice. There is no equivalence between regular CNNs and graph CNNs with polynomial filters. Let us also mention the work of \cite{atwood16} where the convolution is defined using a diffusion process on the graph. Due to lack of space, we do not report the exact definition but this one shares similarities with \eqref{eq:conv_poly} where the normalised transition matrix $\ma{P} = \ma{D}^{-1}\ma{W}$ is substituted for the Laplacian.

In our work, we built upon the work of \cite{niepert16} and \cite{monti16} to get rid of these shortcomings. The convolutions are directly defined in the vertex domain in a way which allows ones to directly control the computational complexity and the localisation of the filters.
Furthermore, these filters do not suffer from the isotropy issue of polynomial filters.

\subsection{Our method}

Each layer of our graph CNN implements a function 
\begin{align}
\label{eq:cnn_function}
f 
\colon 
\left(\Sig, \Graph\right) \quad\;
& \longmapsto 
f\left(\Sig, \Graph \right)
\end{align}
where $\Sig \in \Rbb^{\nbVert \times \nbFeat_0}$ is the input signal, $f\left(\Sig, \Graph \right) \in \Rbb^{\nbVert \times \nbFeat_1}$, and $\Graph$ is a $n$-vertex graph on which the columns of $\Sig$ live and which defines how the convolution is done in this layer. The input signal $\Sig$ has size $\nbVert$ in the ``spatial'' dimension -- \eg, $\nbVert$ pixels for images -- and has $\nbFeat_0$ channels or feature maps -- \eg, $\nbFeat_0~=~3$ for color images. The output signal has same spatial size $\nbVert$ -- we do not use any pooling layers in this work -- and $\nbFeat_1$ feature maps.

\subsubsection{Convolution}

\add{The convolution we use follows principles also used in, \eg, \cite{scarselli09, li16, niepert16, monti16}, where the computation done at one vertex is a function of (at least) the values of the signal at this vertex and neighbouring vertices as well as of labels attributed to each edge. We choose here to use the formalism of \cite{monti16} for our description.}

Convolutions in \cite{monti16} are done in two steps: the extraction of a signal patch around each vertex and a scalar product. We assume here that all vertices have the same number of connections: $\abs{ \{j \colon (i, j) \in \set{E} \} } = \gdeg$ for all $i \in \{1, \ldots, \nbVert \}$. If this is not the case, one can always complete the set of edges and associate to these edges, \eg, a null weight. We also assume that $(i, i) \in \set{E}$ for all $i \in \{1, \ldots, \nbVert\}$.

For a given graph $\Graph$ satisfying the above assumption and with adjacency matrix $\ma{W} \in \Rbb^{\nbVert \times \nbVert}$, we model patch extraction at vertex $i$ with a function 
\begin{align}
p : \{1, \ldots, n\} \times \Rbb^\nbVert & \longrightarrow \Rbb^{\gdeg} \nonumber \\
\left(i, \sigVec \right) \;\; & \longmapsto \left(p_1\left(i, \sigVec \right), \ldots, p_\gdeg\left(i, \sigVec \right)\right)^\adjoint
\end{align}
where each $p_{\ell}\left(i, \sigVec \right) \in \Rbb$, $\ell=1,\ldots,\gdeg$, extracts one entry of the vector $\sigVec$, which represents one column of the input signal $\Sig$. Let $j_1, \ldots, j_\gdeg$ be the $\gdeg$ indices to which $i$ is connected. The order in which these $\gdeg$ entries are extracted by $p$ is determined by ``pseudo-coordinates'' $u(i, j_k) \in \{1, \ldots, \gdeg\}$ attributed to each connected vertex $j_k$ \cite{monti16}. The nature of these pseudo-coordinates will be given in Section~\ref{sec:local_convolution} for local convolution and in Section~\ref{sec:non_local_convolution} for non-local convolution. We define
\begin{align}
\label{eq:pixel_extraction}
p_{\ell}\left(i, \sigVec \right) = g\left( \vec{w}_{i}, j_k \right) \; \sigVec_{j_k}
\end{align}
where $u(i, j_k) = \ell$. The vector $\vec{w}_{i} \in \Rbb^{\nbVert}$ is the $i^\th$ row of $\ma{W}$, which contains at most $\gdeg$ non-zero entries, and $g : \Rbb^\nbVert \times \Rbb \rightarrow \Rbb$ is a re-weighting function that gives the possibility to account for each edge weight in the convolution. We noticed that the choice of this function is very important in the definition of the non-local convolutions to achieve good results in our signal processing applications (see its definition in Section~\ref{sec:non_local_convolution}). Note that $g$ depends on $\vec{w}_{i}$ and $j_k$ in our work while this function depends solely on the pseudo-coordinates in \cite{monti16}. This is a simple but important modification for our applications.

Convoluting $\sigVec$ with a filter $\vec{h} \in \Rbb^\gdeg$ is then defined as in \cite{monti16}:
\begin{align}
\label{eq:convolution}
\left(  \sigVec \star \vec{h}\right) (i) = \vec{h}^\adjoint p(i, \sigVec),
\end{align}
for all $i \in \{1, \ldots, \nbVert \}$. Finally, the function $f$ in \eqref{eq:cnn_function} satisfies
\begin{align}
\label{eq:layer}
f\left(\Sig, \Graph \right)  = \bigg( s \big( \sum_{j=1}^{\nbFeat_0} \sigVec_j \star \vec{h}_j^\ell, \bias^\ell \big) \bigg)_{\ell = 1, \ldots, \nbFeat_1}
\end{align}
where $s : \Rbb \times \Rbb \rightarrow \Rbb$ is an element-wise non-linearity, \eg, $\rm ReLU$ defined as $s\left(\sigVec, \bias \right) = {\rm ReLU}_{\bias}\left( \sigVec \right) = \max \{0, \sigVec + \bias \}$, $\sigVec_j \in \Rbb^{\nbVert}$ denotes the $j^\th$ column-vector of $\Sig$, $\vec{h}_j^\ell \in \Rbb^d$, $j=1, \ldots, \nbFeat_0$, $\ell = 1, \ldots, \nbFeat_1$, are filters, and $\bias^1, \ldots, \bias^{\nbFeat_1}$ are biases. 

Let us highlight that the size $\nbVert$ of the input signal $\Sig$ in the spatial dimension is not fixed in \eqref{eq:layer}. Hence, $f$ can be computed for signals of different sizes using the same filters $\vec{h}_j^\ell$, exactly as with regular CNNs. 

We explain in the next section how one can recover the usual local convolution for images from this definition. We will then continue with the description of the proposed non-local filtering in the Section~\ref{sec:non_local_convolution}.

\subsubsection{Local convolution}
\label{sec:local_convolution}

As noticed in \cite{monti16}, the above definition of convolution permits us to recover easily the standard convolution for images (or, similarly, signals on regular lattices) by constructing a local graph from the Cartesian $2D$-coordinates of each pixel in the image.\footnote{We consider a regular grid of equispaced pixels.} We denote these coordinates $(\alpha(i), \beta(i))$, $i = 1, \ldots, \nbVert$. For a filter of size $\sqrt{\gdeg} \times \sqrt{\gdeg}$, we connect each pixel $i$ to all its local neighbours $j_1, \ldots, j_\gdeg$ that satisfies
$
\abs{\alpha(j_k) - \alpha(i)} \leq \floor{\sqrt{\gdeg}/2}
\text{  and }
\abs{\beta(j_k) - \beta(i)} \leq \floor{\sqrt{\gdeg}/2},
$
%
for $k = 1, \ldots, \gdeg$. We then build the local adjacency matrix $\ma{W}$ that satisfies \
$\ma{W}_{ij} = 1$ if $(i, j) \in \set{E}$, and 0 otherwise. 
%
%
The pseudo-coordinates are determined using the relative position of each pixel $j_k$ to pixel $i$. For any pixel of the image, the connected pixels have relative coordinates in $\{(\alpha(i)-\alpha(j_k), \beta(i)-\beta(j_k)), 1 \leq k \leq \gdeg\}$. We thus create a look up table ${\rm c} : \Rbb \times \Rbb \rightarrow \{1, \ldots, d\}$ that associates a unique integer $\ell \in \{1, \ldots, d\}$ to each of these relative coordinates. Then, we define $u(i,j_k) = c \, (\alpha(i)-\alpha(j_k), \beta(i)-\beta(j_k))$.

With this procedure the pixels are always extracted in the same order, \eg, lexicographically. Finally, \eqref{eq:convolution} is equivalent to the usual convolution when using $g(\cdot) = 1$ in \eqref{eq:pixel_extraction}.

\subsubsection{Non-local convolution}
\label{sec:non_local_convolution}

We now describe our proposition to perform more general non-local convolutions, \ie, we give the definition of the pseudo-coordinates and of the function $g$ in \eqref{eq:pixel_extraction}. In our applications, these convolutions are based on a graph $\Graph$ that captures some structure that we wish to preserve in the signal $\Sig$. The exact construction of $\Graph$ thus differs depending on the application. Yet, we define the pseudo-coordinates and the function $g$ always in the same way, whatever the application. 

The weight $\ma{W}_{ij}$ of the edge between vertices $i$ and $j$ is determined based on a distance between feature vectors $\vec{f}_i$ and $\vec{f}_j$ extracted at vertex $i$ and $j$, respectively. Let $\Delta(\vec{f}_i, \vec{f}_j) \geq 0$ denote this distance. Let again $j_1, \ldots, j_\gdeg$ be the vertices to which vertex $i$ is connected and ordered such that $\Delta(\vec{f}_i, \vec{f}_{j_1}) \leq \ldots \leq \Delta(\vec{f}_i, \vec{f}_{j_\gdeg})$. We propose to define the pseudo-coordinates as $u(i,j_k) = k$, $k \in \{1, \ldots, \gdeg\}$. In other words, the pseudo-coordinates re-order the distances between feature vectors in increasing order. Note that we break any tie arbitrarily. 

Finally, we propose to use the following function $g$ in \eqref{eq:pixel_extraction}:
\begin{align}\label{eq:weights}
g(\vec{w}_i, j) = \frac{\vec{w}_{ij}}{\sum_{k=1}^\nbVert \vec{w}_{ik}} \; \gdeg,
\end{align}
where $\vec{w}_{ij} = \ma{W}_{ij}$ is the $j^\th$ entry of $\vec{w}_{i}$.


%
\section{Style transfer with graph CNNs}
\label{sec:experiments}

In this section, we substitute $f(\Sig)$ for $f(\Sig, \Graph)$ in \eqref{eq:layer} to simplify notations. However, one should not forget that the convolution at each layer is defined by an underlying graph $\Graph$. This graph will always be defined explicitly in the text.

Style transfer consists in transforming a target image $\Sig_t \in \Rbb^{\nbVert \times 3}$, typically a photograph, to give it the ``style'' of a source image $\Sig_s \in \Rbb^{\nbVert ' \times 3}$, typically a painting. Impressive results have recently been obtained using CNNs \cite{gatys16}. The style transfer method of Gatys \emph{et al.} consists in solving a minimization problem of the form
\begin{align}
\label{eq:style_transfer_gatys}
\Sig^* \in \argmin_{\Sig \in \Rbb^{\nbVert \times 3}} 
\sum_{\ell \in \mathcal{L}_s} \norm{f_{\ell}(\Sig)^\adjoint f_{\ell}(\Sig) - f_{\ell}(\Sig_s)^\adjoint f_{\ell}(\Sig_s)}_{\rm F}^2 + \lambda \sum_{\ell \in \mathcal{L}_t}  \norm{f_{\ell}(\Sig) - f_{\ell}(\Sig_t)}_{\rm F}^2,
\end{align}
where $f_{\ell}(\Sig)$ is a matrix with the feature maps at depth $\ell$ of a multi-layer CNN, $\mathcal{L}_s$ and $\mathcal{L}_t$ are two subsets of depths, and $\norm{\cdot}_{\rm F}$ denotes Frobenius norm. Gatys \emph{et al.} used the very deep VGG-19 network, pre-trained for image classification \cite{simonyan14}. The first term encourages the solution $\Sig^*$ to have the style of the painting $\Sig_s$ by matching the Gram matrices of the feature maps, such statistics capturing texture patterns at different scales. The second term ensures that the main structures (the ``content'') of the original photograph $\Sig_t$, as captured in feature maps, are preserved in $\Sig^*$. Note that all the spatial information is lost in the first term that encodes the style, while it is still present in the second term.
It was proved shortly after that similar results can be obtained using a deep neural network with all the filter coefficients chosen randomly \cite{he16}. Let us also mention that \cite{ustyuzhaninov16} showed that texture synthesis, \ie, when only the first term in \eqref{eq:style_transfer_gatys} is involved, can be done using \emph{multiple (8) one-layer} CNNs with random filters giving each $\nbFeat = 1024$ feature maps.

We show now that our graph-based CNNs allow us to revisit neural style transfer. We use only \emph{two one-layer} graph CNNs with \emph{random filters} giving each only $50$ feature maps. This is a much ``lighter'' network than the ones used in the literature. The first network, denoted $f_{1}$, uses local convolutions (Section~\ref{sec:local_convolution}) and the second, denoted $f_{2}$, uses non-local convolutions (Section~\ref{sec:non_local_convolution}) on a graph that captures the structure of the photograph $\Sig_t$ to be preserved. Both $f_{1}$ and $f_{2}$ have the form~\eqref{eq:layer} with $\nbFeat_0 = 3$ for the three \Lab\ channels of color images and $\nbFeat_1 = 50$. We also choose $d=25$ and ReLU for the non-linearity in both cases. The $25 \times 3 \times 50 \times 2 = 7500$ coefficients of the filters $\vec{h}_j^\ell$ and the $50 \times 2 = 100$ biases $\bias^\ell$ in~\eqref{eq:layer} are chosen randomly using independent draws from the standard Gaussian distribution.

The graph in the second CNN $f_{2}$ is constructed as follows. For an image of interest $\Sig$, we construct a feature vector $\vec{f}_i \in \Rbb^{29}$ at each pixel $i$ of the image by extracting all the pixels' \Lab\ values in the neighbourhood of size $3 \times 3$ around $i$ as well as the absolute 2D coordinates of the pixel. We then search the $d=25$ nearest neighbours to $\vec{f}_i$ in the set $\{\vec{f}_1, \ldots, \vec{f}_\nbVert \}$ using the Euclidean distance. Let $\set{D} = \{ \norm{\vec{f}_i - \vec{f}_j}_2 \}_{ij}$ be the set of all distances between each $\vec{f}_i$ and its nearest neighbours. We have $\abs{\set{D}} = 25\nbVert$. To avoid that some pixels are too weakly connected to others, which then produces artefacts in the final images, we compute the $80^\th$ percentile of the values in $\set{D}$ and saturates all the distances above this percentile to this value. The weights of the adjacency matrix $\ma{W}$ then satisfy 
\begin{align}
\ma{W}_{ij} = \exp\left(- \norm{\vec{f}_i - \vec{f}_j}_2^2/\sigma^2 \right),~\forall (i,j)\in\set{E},
\end{align}
with $\sigma$ equal to the $75^\th$ percentile of $\set{D}$. 

\begin{figure}[h!]
\includegraphics[width=0.16\linewidth]{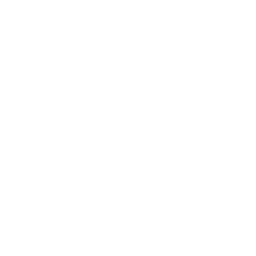}
\includegraphics[width=0.16\linewidth]{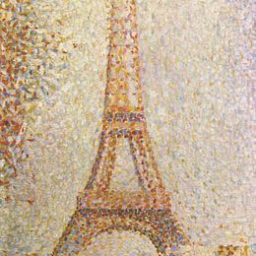}
\includegraphics[width=0.16\linewidth]{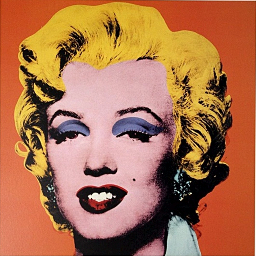}
\includegraphics[width=0.16\linewidth]{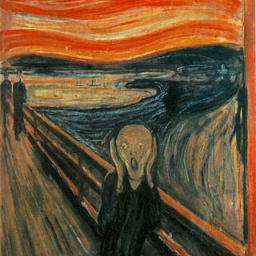}
\includegraphics[width=0.16\linewidth]{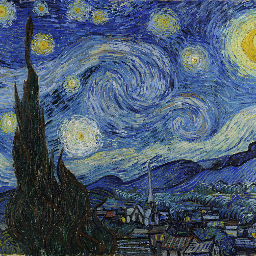}
\includegraphics[width=0.16\linewidth]{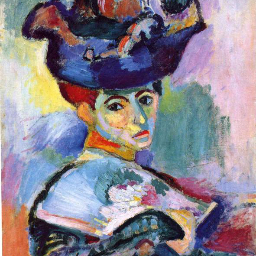}
\includegraphics[width=0.16\linewidth]{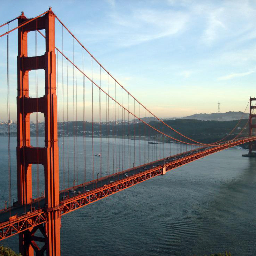}
\includegraphics[width=0.16\linewidth]{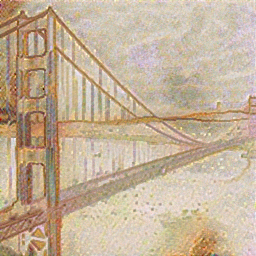}
\includegraphics[width=0.16\linewidth]{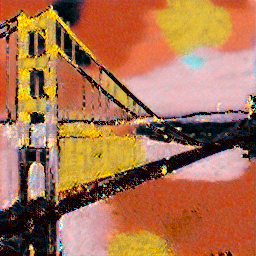}
\includegraphics[width=0.16\linewidth]{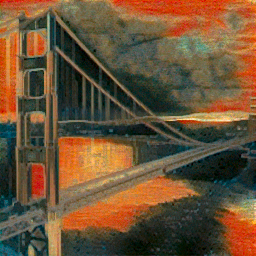}
\includegraphics[width=0.16\linewidth]{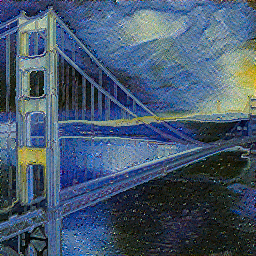}
\includegraphics[width=0.16\linewidth]{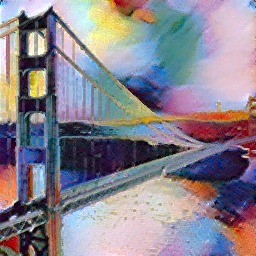}
\includegraphics[width=0.16\linewidth]{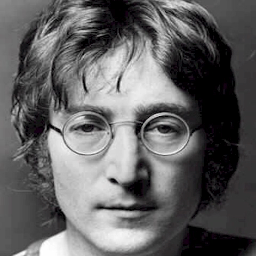}
\includegraphics[width=0.16\linewidth]{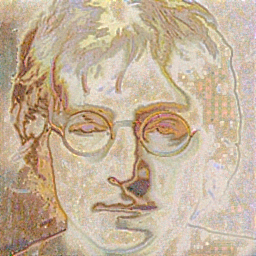}
\includegraphics[width=0.16\linewidth]{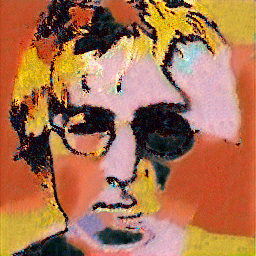}
\includegraphics[width=0.16\linewidth]{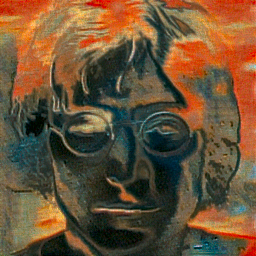}
\includegraphics[width=0.16\linewidth]{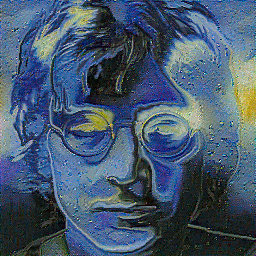}
\includegraphics[width=0.16\linewidth]{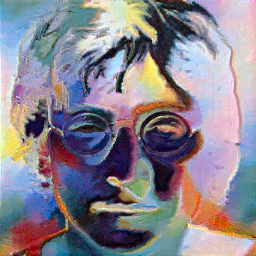}
\includegraphics[width=0.16\linewidth]{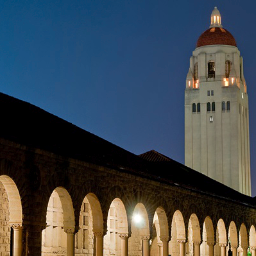}
\includegraphics[width=0.16\linewidth]{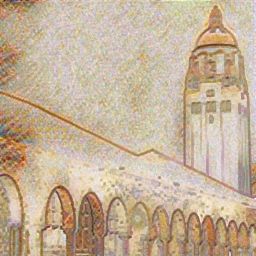}
\includegraphics[width=0.16\linewidth]{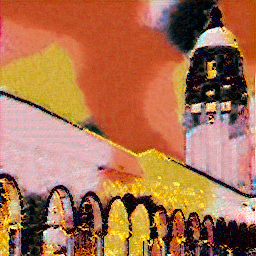}
\includegraphics[width=0.16\linewidth]{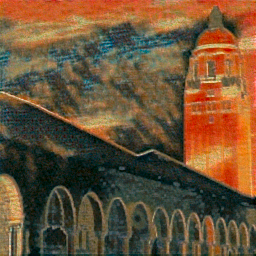}
\includegraphics[width=0.16\linewidth]{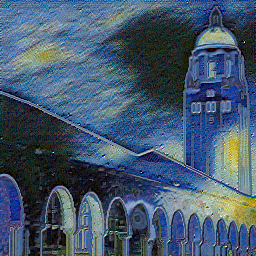}
\includegraphics[width=0.16\linewidth]{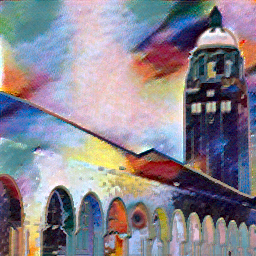}
\includegraphics[width=0.16\linewidth]{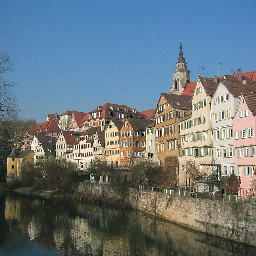}
\includegraphics[width=0.16\linewidth]{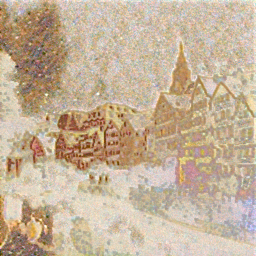}
\includegraphics[width=0.16\linewidth]{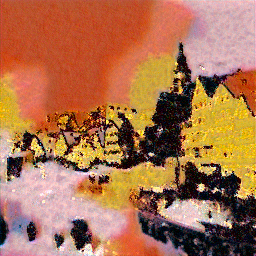}
\includegraphics[width=0.16\linewidth]{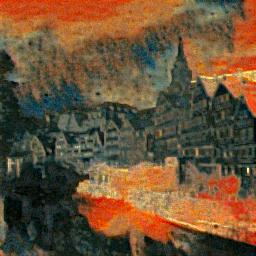}
\includegraphics[width=0.16\linewidth]{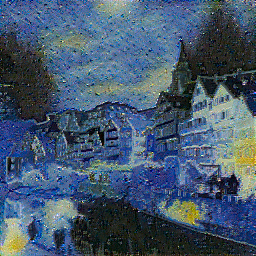}
\includegraphics[width=0.16\linewidth]{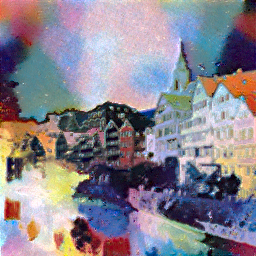}
\caption{\label{fig:style_transfer}Examples of style transfer results obtained where the photograph (left) is transformed to have the style of a given painting (top). Using the proposed graph CNN framework, only two one-layer random CNNs are required to extract matched statistics.}
\end{figure}

We capture the style of the painting $\Sig_s$ by computing the Gram matrices 
\begin{align}
\ma{G}_1 = f_1(\Sig_s)^\adjoint f_1(\Sig_s)
\text{ and }
\ma{G}_2 = f_2(\Sig_s)^\adjoint f_2(\Sig_s), 
\end{align}
where the non-local convolution in $f_2$ is computed using the graph constructed on $\Sig_s$. The matrix $\ma{G}_1$ captures local statistics while $\ma{G}_2$ captures non-local statistics. To give the style of $\Sig_s$ to $\Sig_t$, we now compute a new graph on $\Sig_t$. We then compute an image $\Sig^* \in \Rbb^{\nbVert\times 3}$ by solving
\begin{align}
\label{eq:style_transfer_cnn}
\min_{\Sig \in \Rbb^{\nbVert\times 3}} 
\gamma_1 \norm{f_1(\Sig)^\adjoint f_1(\Sig) - \ma{G}_1}_{\rm F}^2
+ \gamma_2 \norm{f_2(\Sig)^\adjoint f_2(\Sig) - \ma{G}_2}_{\rm F}^2 
+ \gamma_3 \norm{\Sig}_{\rm TV},
\end{align}
where $f_2$ uses, this time, \emph{the graph constructed on $\Sig_t$} for the non-local convolutions, $\norm{\cdot}_{\rm TV}$ is the Total Variation norm, and $\gamma_1, \gamma_2, \gamma_3 > 0$. Note that unlike in~\eqref{eq:style_transfer_gatys}, we do not try to match feature maps but only Gram matrices. Yet the final image $\Sig^*$ retains the structure of $\Sig_t$ thanks to the non-local convolution in $f_2$.

In practice, we minimise \eqref{eq:style_transfer_cnn} using the L-BFGS algorithm starting from a random initialisation of $\Sig$. The parameters $\gamma_1, \gamma_2$ are computed so that the gradient coming from the term they respectively influence has a maximum amplitude of $1$ at the first iteration of the algorithm. We set $\gamma_3 = 0.01\nbVert$. All images used in the experiments have size $\nbVert = 256 \times 256$. However, we do not solve~\eqref{eq:style_transfer_cnn} directly at this resolution, but in a coarse-to-fine scheme instead: We start by downsampling all images at $\nbVert = 32 \times 32$ pixels; Solve~\eqref{eq:style_transfer_cnn} at this resolution; Upscale the solution at $64 \times 64$ pixels; Restart the same process at this new resolution using the up-scaled image as initialisation; Repeat this process until the final resolution is reached.

We present some results obtained with our graph-based method in Fig.~\ref{fig:style_transfer}. One can notice that the main structure of the photograph $\Sig_t$ perfectly appears in $\Sig^*$ thanks to the presence of the structure-preserving non-local convolutions in $f_2$. The style is also well transferred thanks to the matching of the Gram matrices $\ma{G}_1$ and $\ma{G}_2$.

\begin{figure*}
\includegraphics[width=0.16\linewidth]{styles/blank}
\includegraphics[width=0.16\linewidth]{styles/eiffel}
\includegraphics[width=0.16\linewidth]{styles/marilyn}
\includegraphics[width=0.16\linewidth]{styles/scream}
\includegraphics[width=0.16\linewidth]{styles/starry}
\includegraphics[width=0.16\linewidth]{styles/woman}
\includegraphics[width=0.16\linewidth]{photos/bridge}
\includegraphics[width=0.16\linewidth]{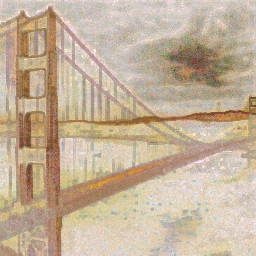}
\includegraphics[width=0.16\linewidth]{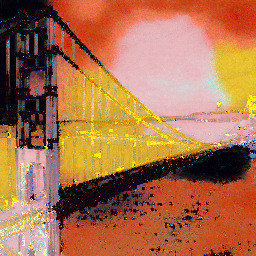}
\includegraphics[width=0.16\linewidth]{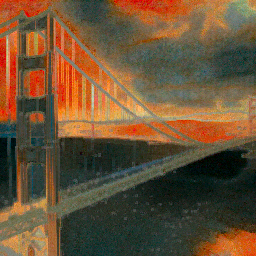}
\includegraphics[width=0.16\linewidth]{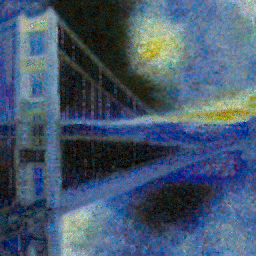}
\includegraphics[width=0.16\linewidth]{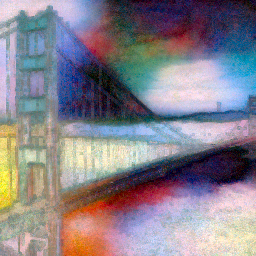}
\caption{\label{fig:style_gcnn}Examples of style transfer results obtained with our graph CNN method using \emph{only non-local convolutions} where the photograph (left) is transformed to have the style of a given painting (top).}
\end{figure*}

\add{To highlight the role of the graph CNN with non-local convolutions, we repeat exactly the same experiments but using only the non-local graph CNN, \ie, we do not use the regular CNN with local convolutions -- the TV regularisation is still present. Fig.~\ref{fig:style_gcnn} shows results for one photograph and different paintings. First, we notice that the main structures of the photograph are well preserved thanks to the graph CNN. Second, the colors of the painting and the relative arrangement of the colors are well transferred. However, we are not able to transfer finer style details like brush strokes. On the contrary, the local CNN is able to capture these finer details which appear in the results with the complete method.} 

\add{Let us highlight that this experiment already shows that our graph CNN framework can adapt to many changes in the graph structure. Indeed, the graph used in $f_2$ was built from the painting when computing $\ma{G}_2$ while it is built from the photograph when computing $\ma{X}^*$.}

%

%
\section{Conclusion}

We proposed a graph CNN framework that allows us to unify local and non-local processing of signals on graphs, \add{and showed how to use this framework to perform style transfer. The results already suggest that the proposed convolution adapts correctly to changes in the input graph. Additional experiments in \suppl\ demonstrate the versatility of our framework on other kinds of signals and tasks.} Beyond signal processing, we believe that some of the tools presented here can be useful to other applications involving time-varying graph structures, such as in social networks.

%
\titleformat{\section}[hang]{\color{blue1}\large\bfseries}{\thesection}{0mm}{}[]
\bibliographystyle{IEEEtran}
\bibliography{biblio}

%
\titleformat{\section}[hang]{\color{blue1}\large\bfseries\centering}{Appendix \thesection}{0mm}{}[]
\appendix
%
\section{ - Color transfer}

\add{In this appendix, we address another application that is color transfer where the goal is to transfer the ``color palette'' of a source image onto the one of the target image. First, we describe how we perform this task using graph CNNs. Second, we describe a more traditional approach using optimal transport, such as used in \cite{frigo14}, with which we will compare our results.}

\subsection{Color transfer with graph CNNs}

All images are converted to \Lab\ and only chrominances \textit{a} and \textit{b} are processed -- the luminance of target images remains untouched. 
The first step consists in building a palette representative of the color distribution of each image. This is done by clustering image's chrominances into $\nbVert = 64$ clusters using $k$-means.
We thus obtain a source palette $\Sig_s \in \Rbb^{\nbVert \times 2}$ and a target one $\Sig_t \in \Rbb^{\nbVert \times 2}$. The first and second columns of these palettes represent respectively the $\textit{a}$ and $\textit{b}$ channels. The pixel values in the \ab\ channels are normalised and shifted to be in the range $[-0.5, 0.5]$.

Similarly to what was done for style transfer, we capture the ``statistics'' of the reference palette $\Sig_s$ by computing the Gram matrix of the output feature maps of a graph convolutional layer $f$ of the form of~\eqref{eq:layer}. We have $\nbFeat_0 = 2$. We choose $\nbFeat_1 = 100$ and $d=10$. The coefficients of the filters $\vec{h}_j^\ell$ and the biases $\bias^\ell$ in~\eqref{eq:layer} are chosen randomly using independent draws from the standard Gaussian distribution. We use ReLU for the non-linearity.

The network $f$ implements non-local convolutions (Section~\ref{sec:non_local_convolution}) with a graph constructed as follows. For a palette $\Sig \in \Rbb^{\nbVert \times 2}$, we construct a feature vector $\vec{f}_i \in \Rbb^{2}$ for each palette element $i$ that simply contains the \ab\ values for that element. We then search the $d= 10$ nearest neighbours to $\vec{f}_i$ in the set $\{\vec{f}_1, \ldots, \vec{f}_\nbVert \}$ using the Euclidean distance. The weights of the adjacency matrix $\ma{W}$ representing the graph $\Graph$ used for convolution satisfy 
\begin{align}
\ma{W}_{ij} = \exp\left(- \norm{\vec{f}_i - \vec{f}_j}_2^2/\sigma^2 \right),
\end{align}
with $\sigma=0.25$, for connected palette entries $i$ and $j$.

We capture the color statistics of the source image by computing the Gram matrix
\begin{align}
\ma{G}_s = f(\Sig_s)^\adjoint f(\Sig_s),
\end{align}
where the graph used 
in $f$ was built using $\Sig_s$.

To find a mapping between the colors in the source and target images, we solve the following minimisation problem
\begin{align}
\label{eq:color_transfer_cnn}
\min_{\Sig \in \Rbb^{\nbVert \times 2}} 
\gamma_1 \norm{f(\Sig)^\adjoint f(\Sig) - \ma{G}_s}_{\rm F}^2 + \gamma_2 \norm{\Sig - \Sig_t}_{\rm F} 
+ \gamma_3 \; {\rm Tr} \left( \Sig^\adjoint \Lap \Sig \right),
\end{align}
where the graph used for the convolutions in $f$ is now built using $\Sig_t$, $\Lap \in \Rbb^{\nbVert \times \nbVert}$ and $\gamma_1, \gamma_2, \gamma_3 > 0$. We explain the role of each term and give the definition of $\Lap$ in the next paragraph. We solve this problem using the L-BFGS algorithm starting from $\Sig_t$ as initialisation. The parameter $\gamma_1$ is computed so that the gradient coming from the term it influences has a maximum amplitude of $1$ at the first iteration of the algorithm. We set $\gamma_2 = 0.1$ and $\gamma_3 = 1/\max_i\{\ma{L}_{ii}\}$. Let $\Sig^* \in \Rbb^{\nbVert \times 2}$ be the obtained solution to~\eqref{eq:color_transfer_cnn}. We transform the color in target image as follows. For each pixel  \ab\ value in this image, we find its nearest neighbour in the palette $\Sig_t$, say the $i$-th color in the palette. The new color is then obtained by replacing the pixel \ab\ value by the $i^\th$ entry of the new palette $\Sig^*$, so that there are only $64$ distinct colors in the final images.

The first term in~\eqref{eq:color_transfer_cnn} ensures that the colors in $\Sig^*$ are similar to the colors in $\Sig_s$. The second and third terms in~\eqref{eq:color_transfer_cnn} permit us to ensure a consistency in the mapping from the old colors $\Sig_t$ to the new colors $\Sig^*$. The second term controls the total cost of moving from the old colors to the new colors. Note that this cost is also usually involved in optimal transport methods for color transfer \cite{ferradans13, frigo14}. The role of the third term is to ensure that two similar palette elements in $\Sig_t$ should map to similar palette elements in $\Sig^*$. The matrix $\Lap$ is the combinatorial Laplacian matrix constructed from a symmetric version $\widetilde{\ma{W}}$ of the adjacency matrix $\ma{W}$ built from $\Sig_t$: $\widetilde{\ma{W}} = (\ma{W} + \ma{W}^\adjoint)/2$. 
The third term in~\eqref{eq:color_transfer_cnn} classically promotes smoothness on the weighted symmetrized graph since
$
{\rm Tr} \left( \Sig^\adjoint \Lap \Sig \right) = \frac{1}{2} \sum_{i,j = 1}^\nbVert \sum_{k=1}^2 \widetilde{\ma{W}}_{ij} \left(\Sig_{ik} - \Sig_{jk} \right)^2.$
%
Similar elements in palette $\Sig_t$ should remain similar in transformed palette $\Sig^*$.

\subsection{Optimal transport}

Let $\ma{C} \in \Rbb^{\nbVert \times \nbVert}$ be the cost matrix with entries $\ma{C}_{ij} = \norm{ (\Sig_t)_i - (\Sig_s)_j }_2$, where $(\Sig_t)_i \in \Rbb^2$ and $(\Sig_s)_j \in \Rbb^2$ are the $i^\th$ and $j^\th$ rows of $\Sig_t$ and $\Sig_s$, respectively. This matrix encodes the cost of moving the palette elements in $\Sig_t$ to the palette elements in $\Sig_s$, and vice-versa. The optimal transport problem is about finding the transport from $\Sig_t$ to $\Sig_s$ that is the least costly:
\begin{align}
\min_{\ma{\Gamma} \in \Rbb^{\nbVert \times \nbVert}} \scp{\ma{C}}{\ma{\Gamma}}_{\mathrm{F}}
\text{ s.t. }
\left\{
\begin{array}{ll}
0 \leq \vec{1}\ma{\Gamma} \leq \nbVert^{-1}, & \vec{1} \ma{\Gamma} \vec{1}^\adjoint = 1, \\
0 \leq \ma{\Gamma}\vec{1}^\adjoint \leq \nbVert^{-1}, & \ma{\Gamma} \geq 0,
\end{array}\right.
\end{align}
where $\vec{1} = (1, \ldots, 1)^\adjoint \in \Rbb^{\nbVert}$ and the inequalities on the right hand side hold element-wise. This is the optimal transport problem solved in \cite{frigo14} for color transfer, except that the cost matrix $\ma{C}$ also incorporates information about the luminance of each image in their work, which hence yields different results. Let $\ma{\Gamma^*} \in \Rbb^{\nbVert \times \nbVert}$ be the solution to the above convex problem. We compute the new palette $\Sig_{ot} \in \Rbb^{\nbVert \times 2}$ whose rows read
\begin{align}
(\Sig_{ot})_{i} = \frac{\sum_{j=1}^\nbVert \ma{\Gamma}_{ij}^* (\Sig_s)_{j}}{\sum_{j=1}^\nbVert \ma{\Gamma}_{ij}^*}
\end{align}
for $i=1, \ldots, \nbVert$. Remark that the palette $\Sig_{ot}$ is made of colors similar to those in the palette $\Sig_s$ as each palette element of $\Sig_{ot}$ is a convex combination of palette elements in $\Sig_s$. We finally transform the color in image $t$ as follows. For each pixel \ab\ value in image $t$, we find the nearest neighbour in the palette $\Sig_t$, say the $i^\th$ entry of $\Sig_t$. The new color is then obtained by replacing the pixel \ab\ value by the $i^\th$ entry of the new palette $\Sig_{ot}$.

\subsection{Results}

\begin{figure*}
\centering
\begin{minipage}{0.24\linewidth}
\centering
Target image
\end{minipage}
\begin{minipage}{0.24\linewidth}
\centering
Our method
\end{minipage}
\begin{minipage}{0.24\linewidth}
\centering
Optimal transport
\end{minipage}
\begin{minipage}{0.24\linewidth}
\centering
Source colors
\end{minipage}\\
\includegraphics[width=0.24\linewidth]{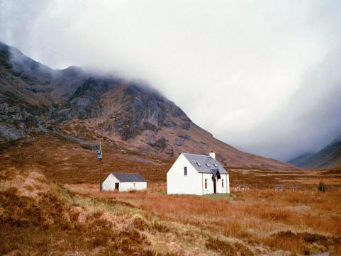}
\includegraphics[width=0.24\linewidth]{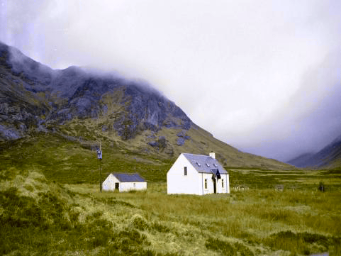}
\includegraphics[width=0.24\linewidth]{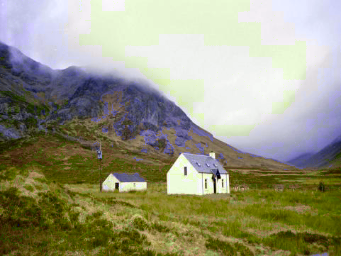}
\includegraphics[width=0.24\linewidth]{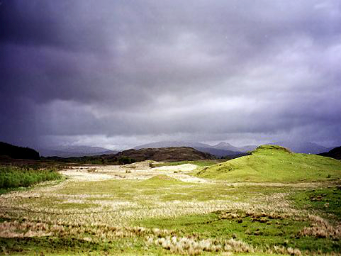}\\
\includegraphics[width=0.24\linewidth]{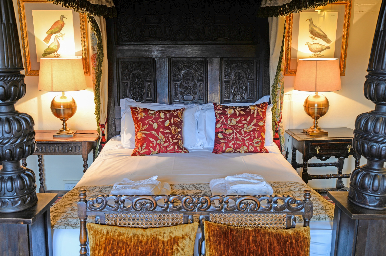}
\includegraphics[width=0.24\linewidth]{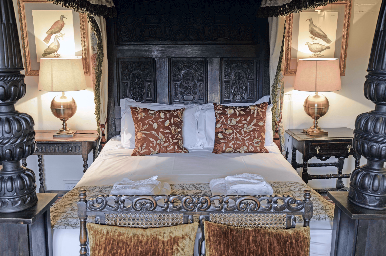}
\includegraphics[width=0.24\linewidth]{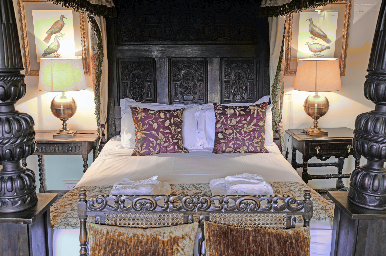}
\includegraphics[width=0.24\linewidth]{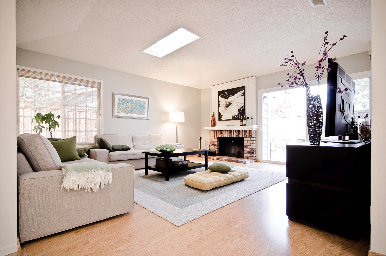}\\
\includegraphics[width=0.24\linewidth]{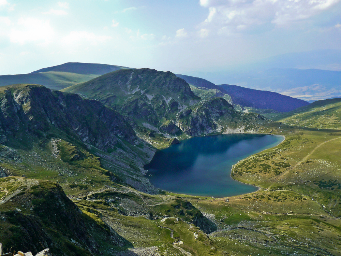}
\includegraphics[width=0.24\linewidth]{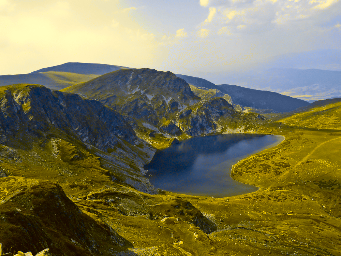}
\includegraphics[width=0.24\linewidth]{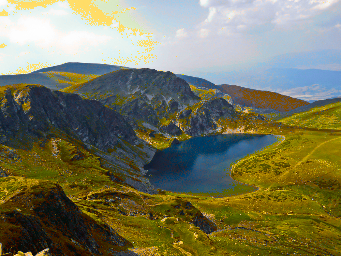}
\includegraphics[width=0.24\linewidth]{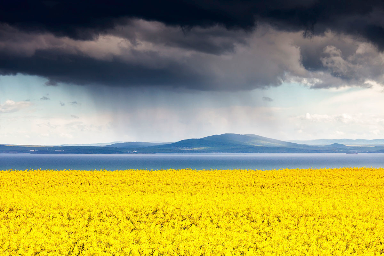}\\
\includegraphics[width=0.24\linewidth]{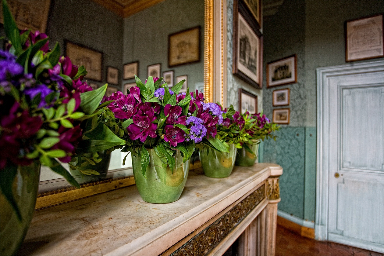}
\includegraphics[width=0.24\linewidth]{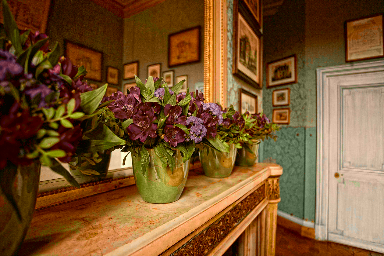}
\includegraphics[width=0.24\linewidth]{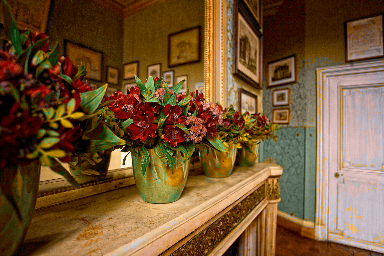}
\includegraphics[width=0.24\linewidth]{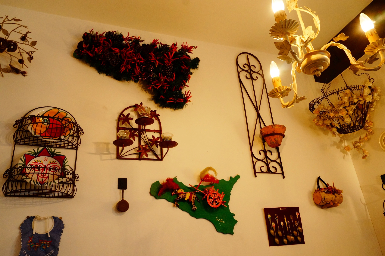}\\
\includegraphics[width=0.24\linewidth]{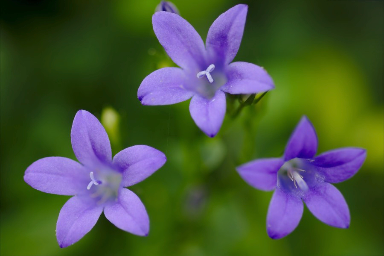}
\includegraphics[width=0.24\linewidth]{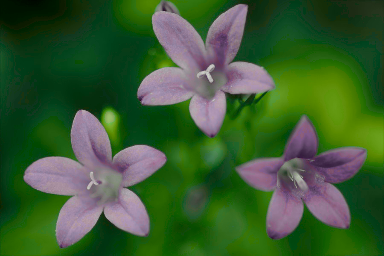}
\includegraphics[width=0.24\linewidth]{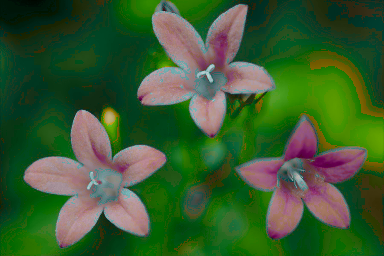}
\includegraphics[width=0.24\linewidth]{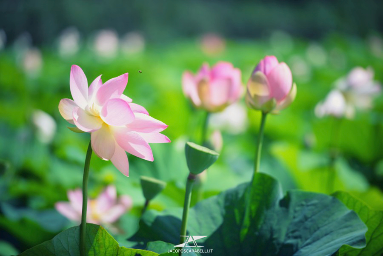}
\caption{\label{fig:color_transfer}Examples of color transfer results where the photograph on the left is modified to take the color of the photograph on the right. The second image from the left shows the result obtained with our graph CNN method while the second from the right shows the result obtained with optimal transport.}
\end{figure*}

We present color transfer results obtained with both methods in Fig.~\ref{fig:color_transfer}. One can notice that our results suffer from fewer artefacts than the ones obtained with optimal transport. One can also refer to the results of \cite{hristova15} on the same images\footnote{\url{http://people.irisa.fr/Hristina.Hristova/publications/2015_EXPRESSIVE/indexResults.html}} for comparison with a method more evolved than a sole optimal transport. We believe that our results achieve similar visual qualities. Let us also mention that the optimal transport results presented here can certainly be improved thanks to some extra graph-regularisation terms such as used in \cite{ferradans13}. Nevertheless, our results show that one can achieve competitive results compared to the state-of-the-art with a completely different approach that uses shallow graph CNNs with random weights.

\begin{figure*}
\centering
\begin{minipage}{0.24\linewidth}
\centering
Target image
\end{minipage}
\begin{minipage}{0.24\linewidth}
\centering
Our method
\end{minipage}
\begin{minipage}{0.24\linewidth}
\centering
Optimal transport
\end{minipage}
\begin{minipage}{0.24\linewidth}
\centering
Image with source colors
\end{minipage}\\
\includegraphics[width=0.24\linewidth]{color_transfer/house}
\includegraphics[width=0.24\linewidth]{color_transfer/house_to_plain_gcnn_0}
\includegraphics[width=0.24\linewidth]{color_transfer/house_to_plain_ot}
\includegraphics[width=0.24\linewidth]{color_transfer/plain}\\
\includegraphics[width=0.24\linewidth]{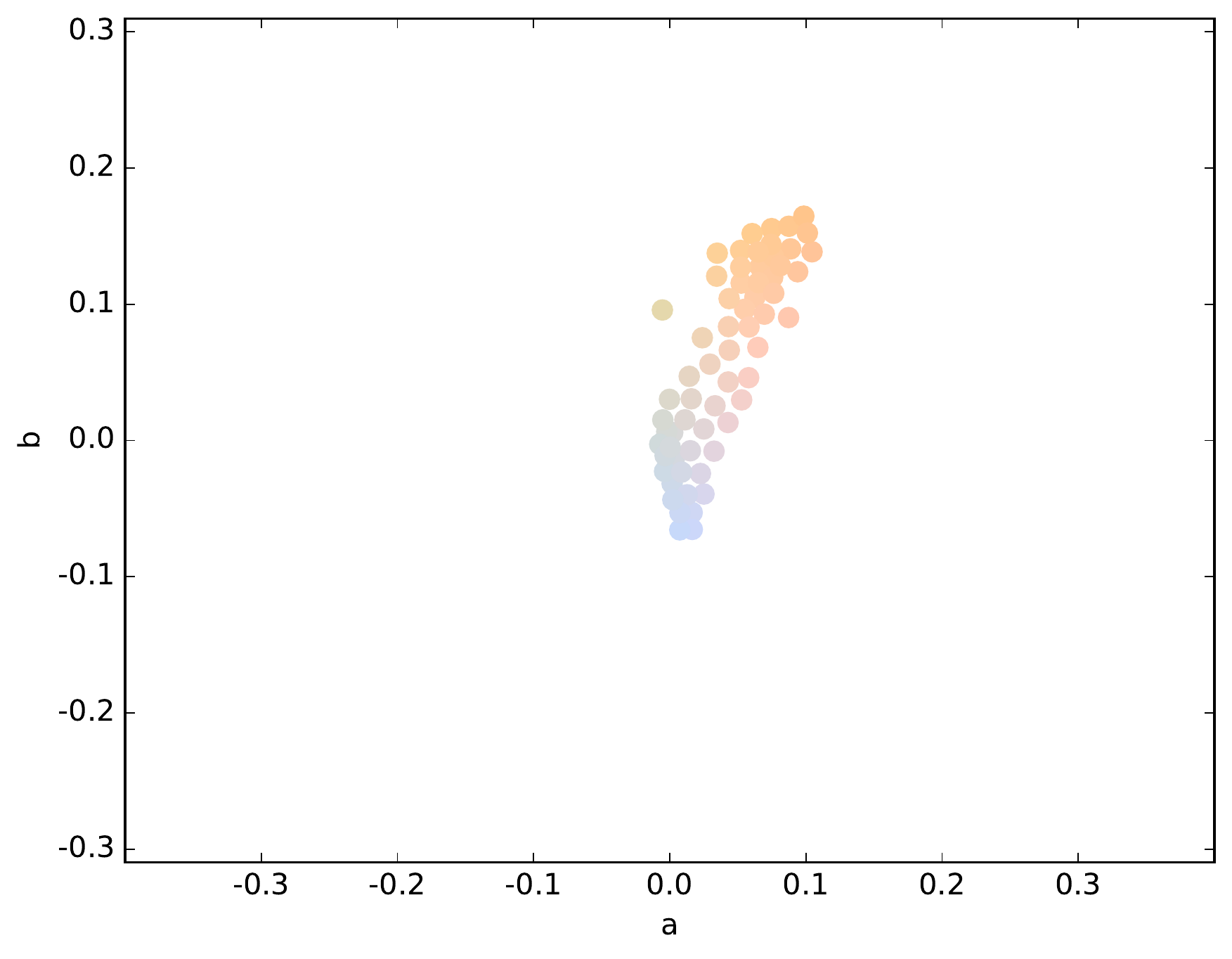}
\includegraphics[width=0.24\linewidth]{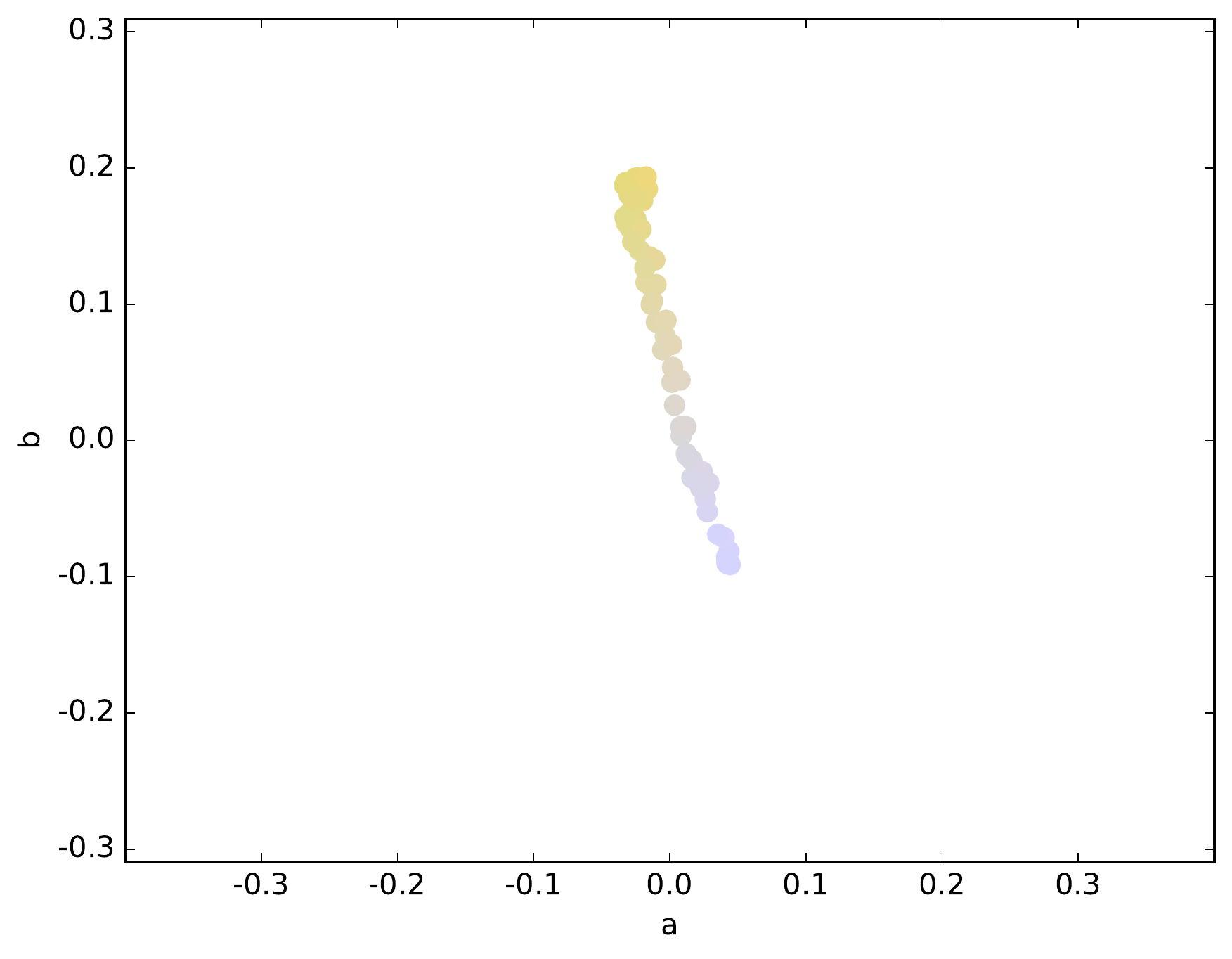}
\includegraphics[width=0.24\linewidth]{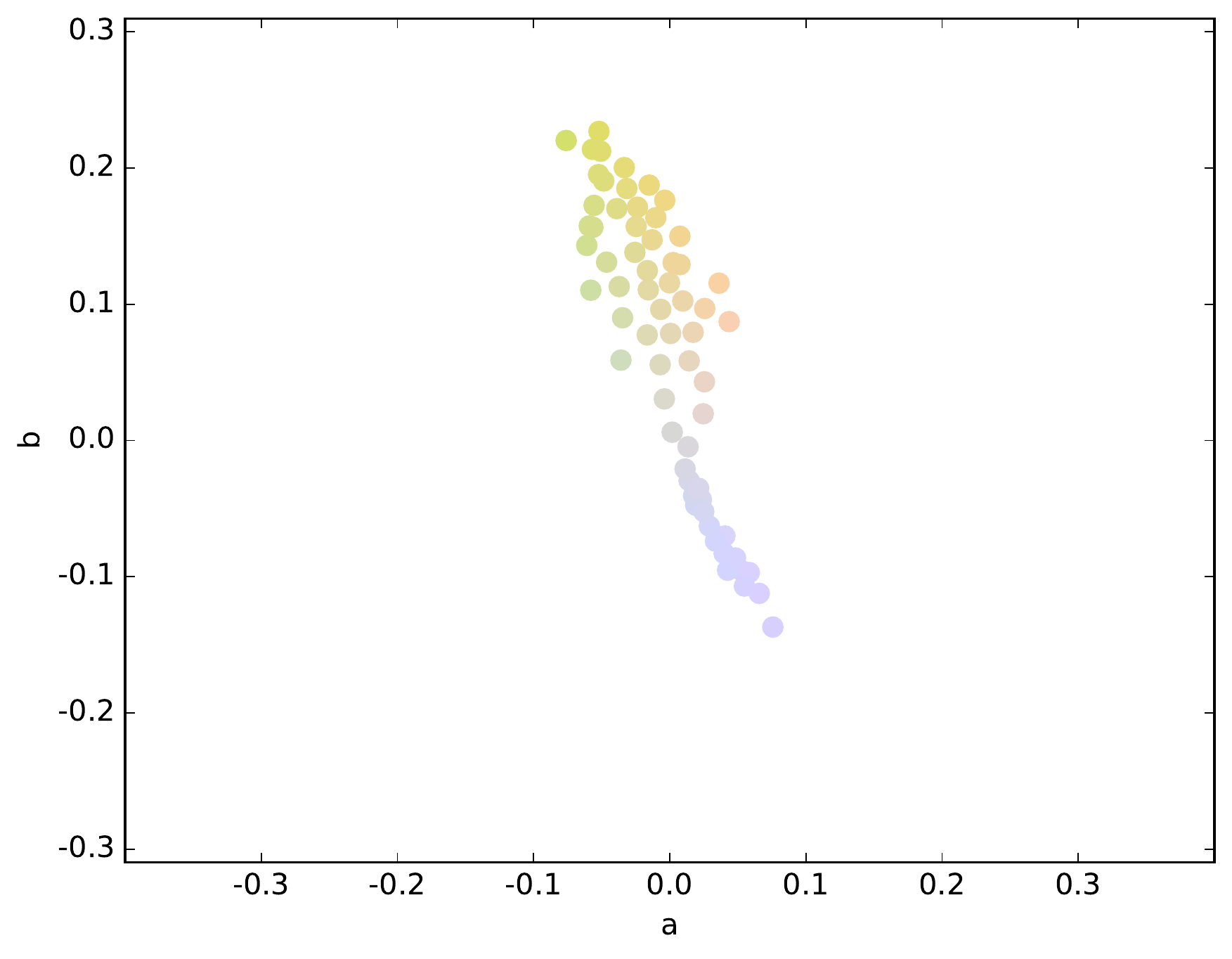}
\includegraphics[width=0.24\linewidth]{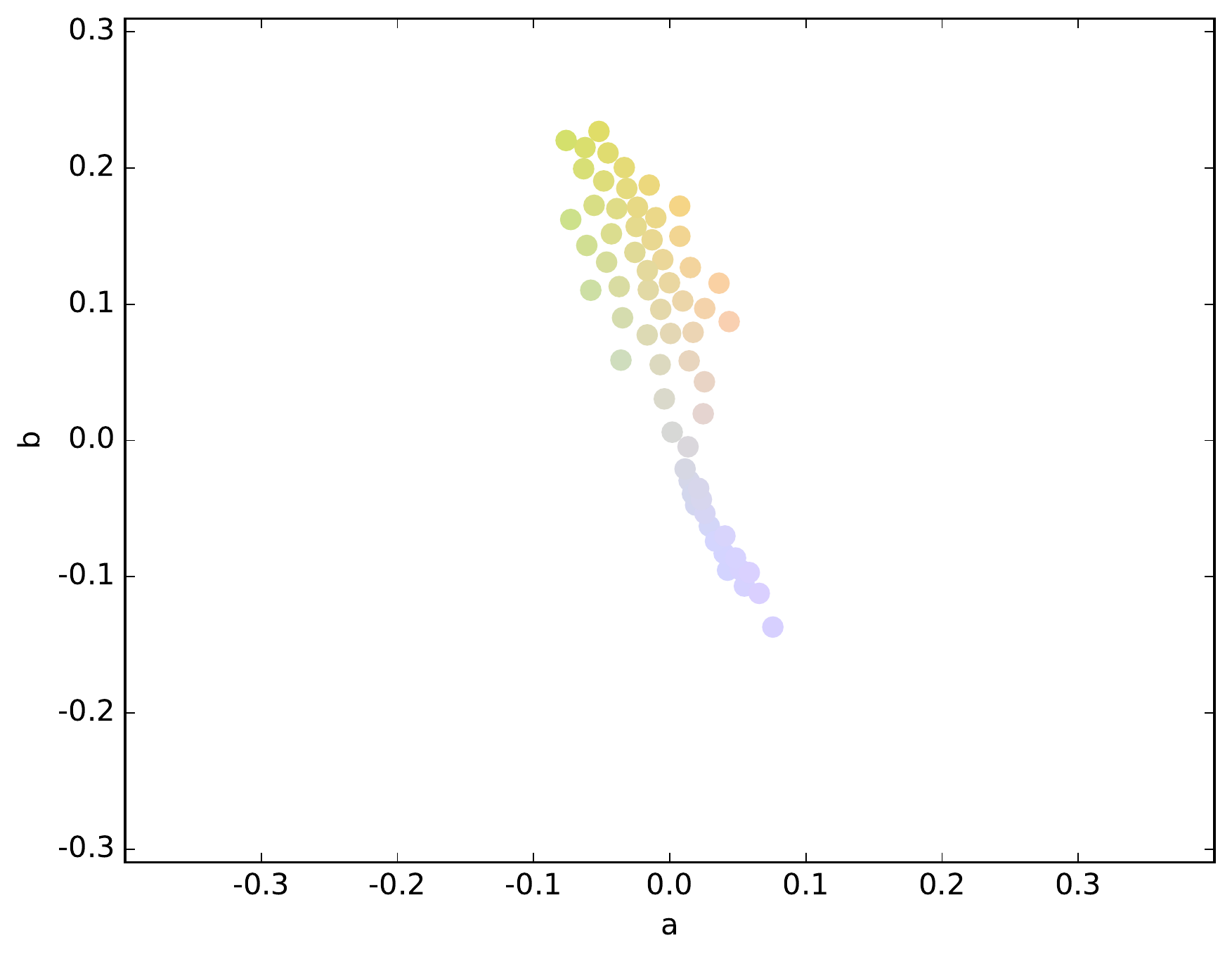}\\
\includegraphics[width=0.24\linewidth]{color_transfer/bedroom}
\includegraphics[width=0.24\linewidth]{color_transfer/bedroom_to_home_gcnn_0}
\includegraphics[width=0.24\linewidth]{color_transfer/bedroom_to_home_ot}
\includegraphics[width=0.24\linewidth]{color_transfer/home}\\
\includegraphics[width=0.24\linewidth]{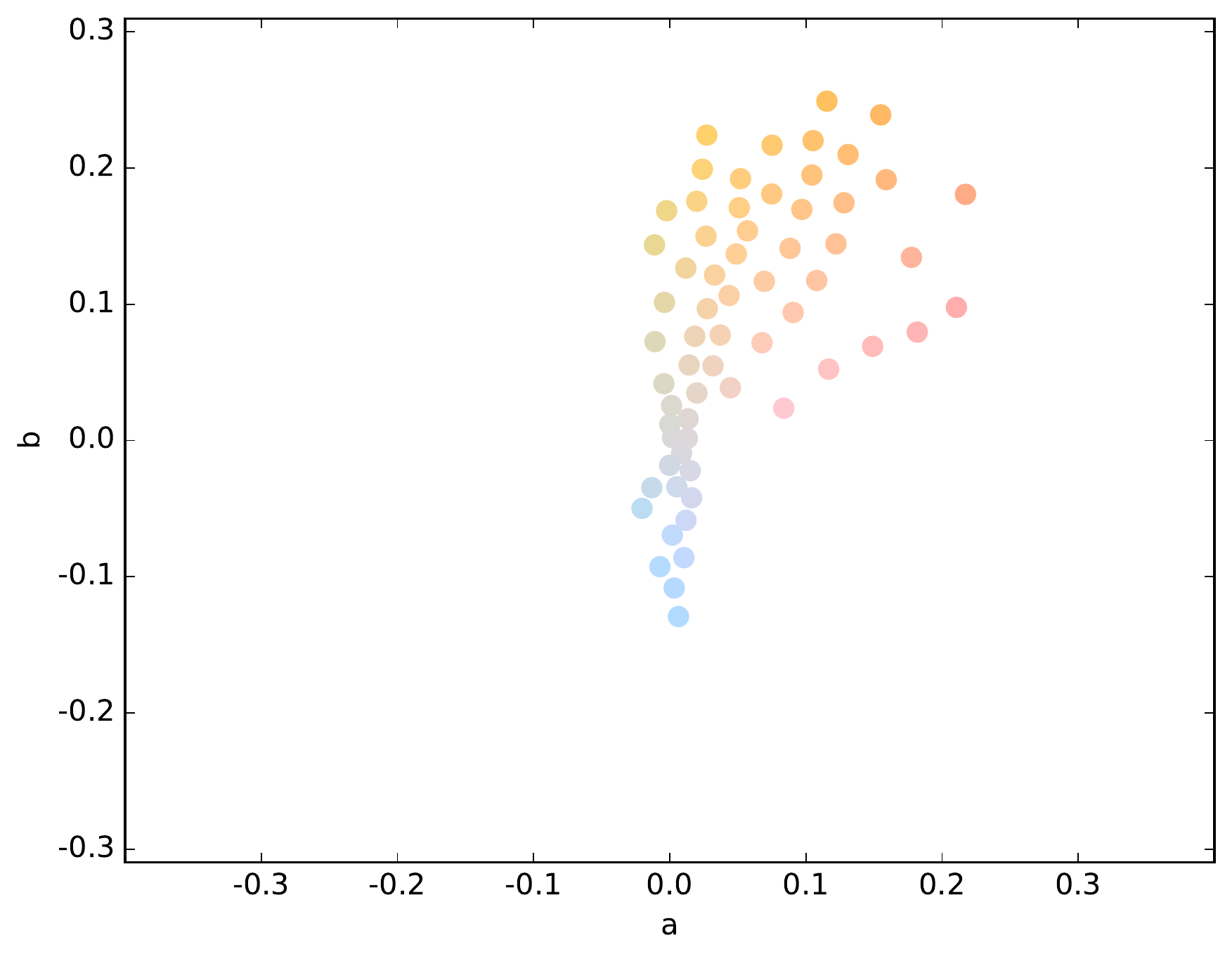}
\includegraphics[width=0.24\linewidth]{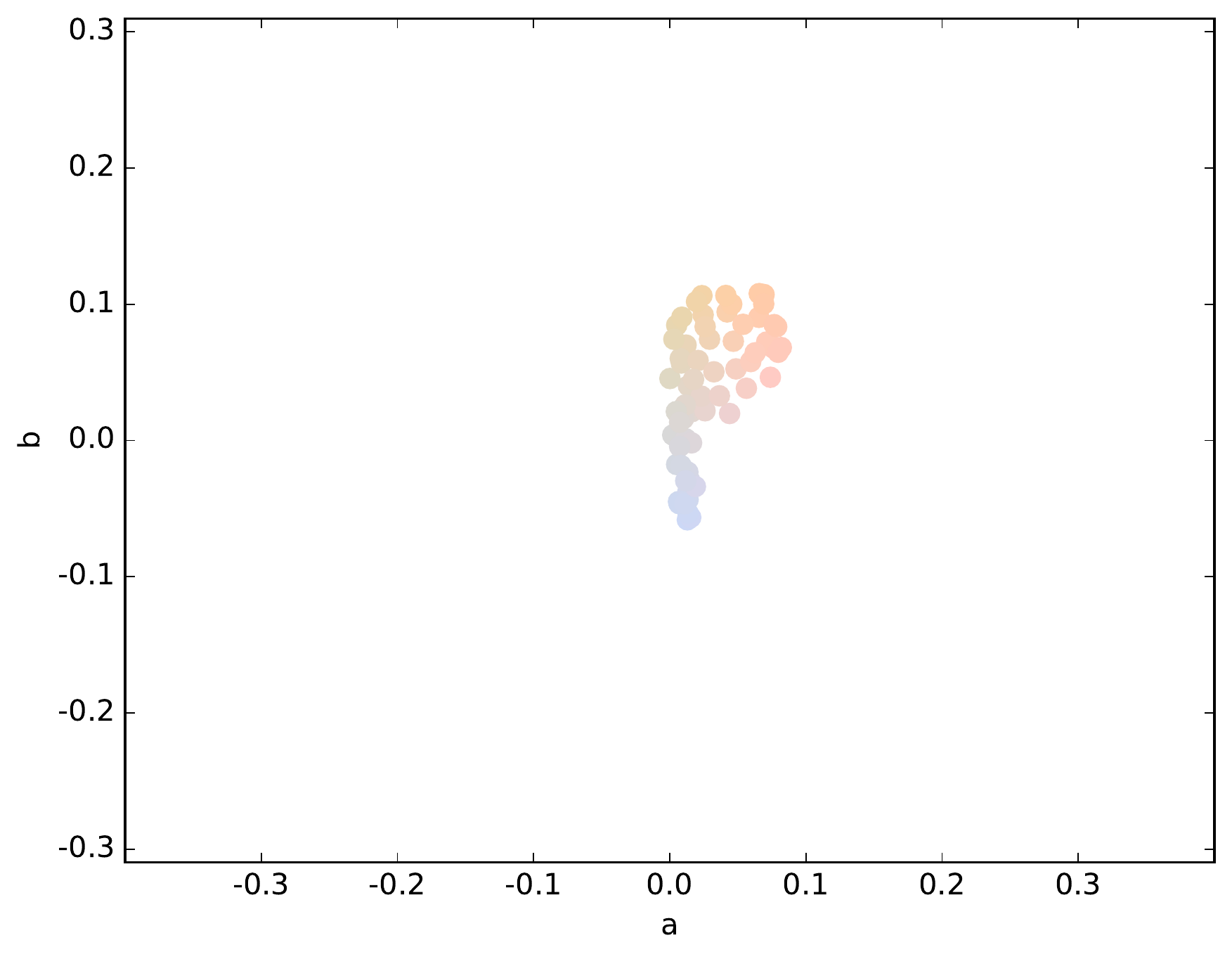}
\includegraphics[width=0.24\linewidth]{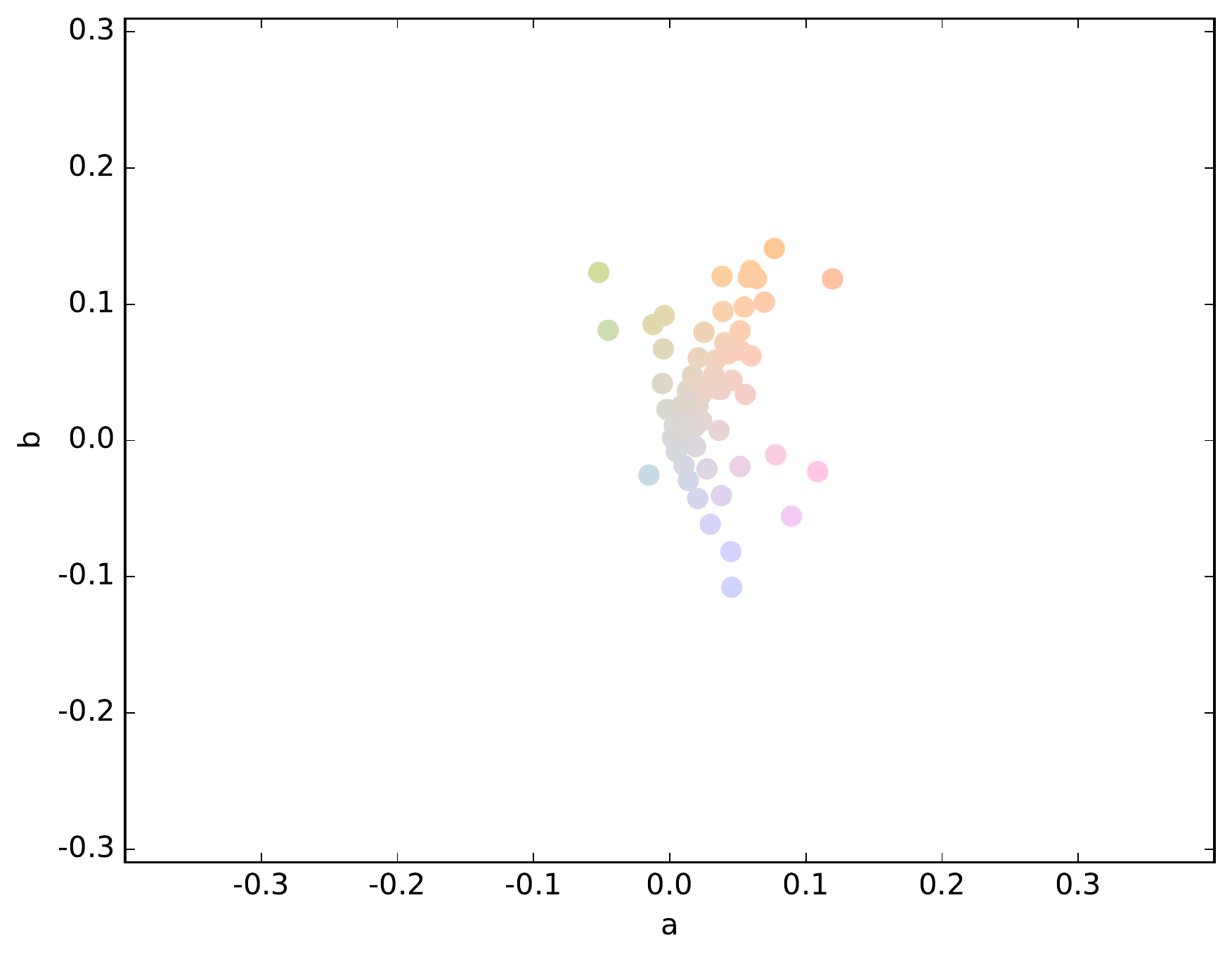}
\includegraphics[width=0.24\linewidth]{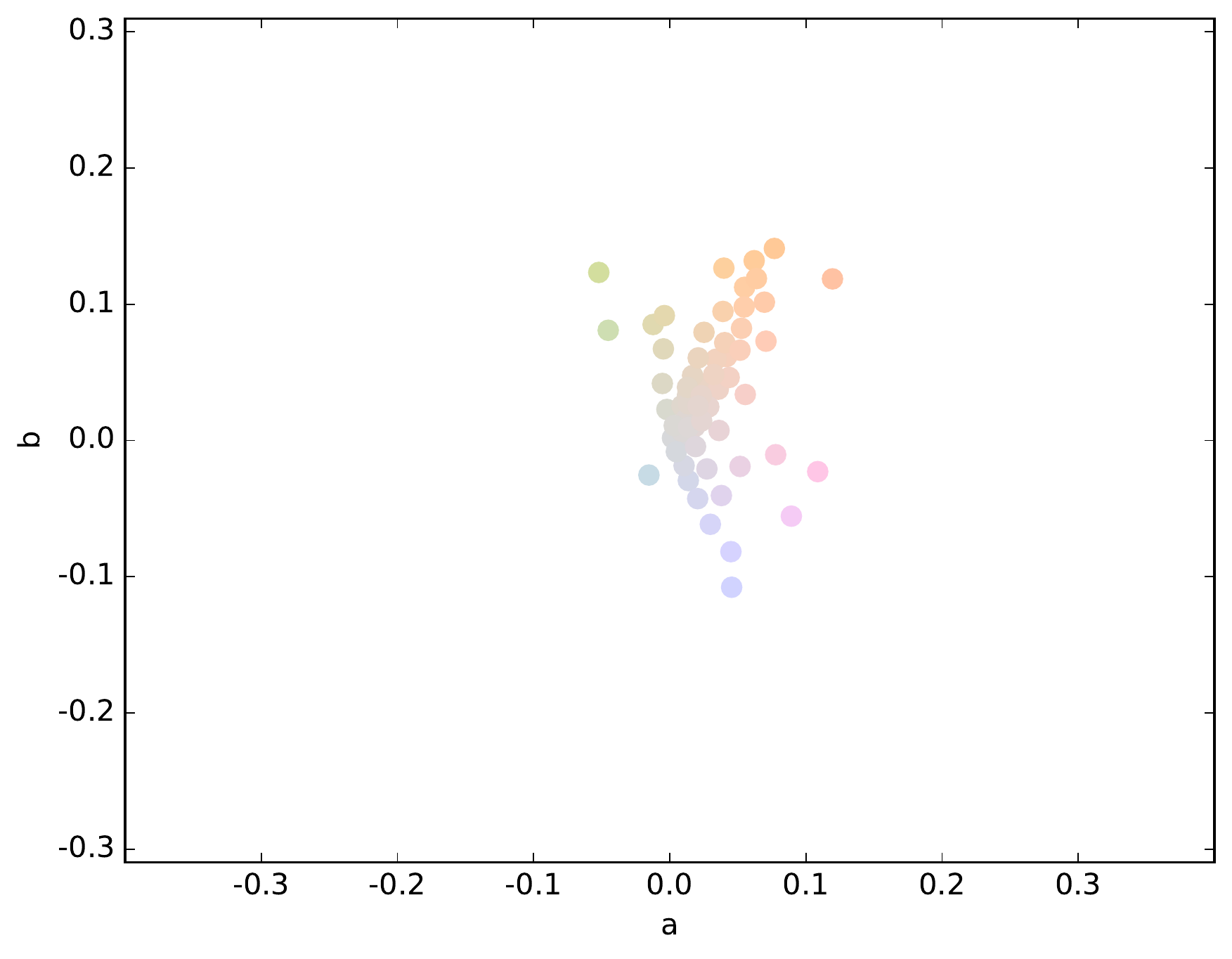}\\
\includegraphics[width=0.24\linewidth]{color_transfer/mountain}
\includegraphics[width=0.24\linewidth]{color_transfer/mountain_to_landscape_gcnn_0}
\includegraphics[width=0.24\linewidth]{color_transfer/mountain_to_landscape_ot}
\includegraphics[width=0.24\linewidth]{color_transfer/landscape}\\
\includegraphics[width=0.24\linewidth]{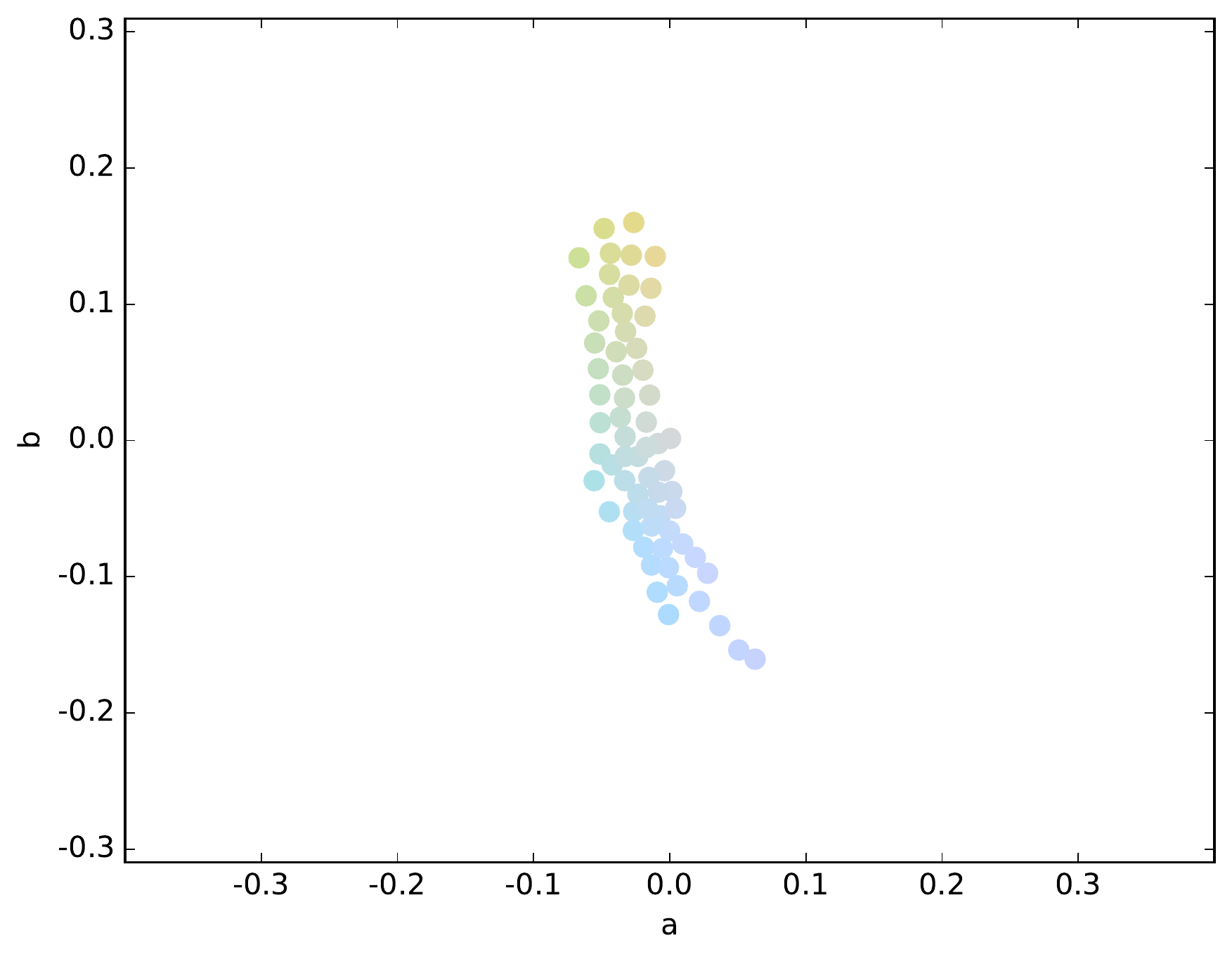}
\includegraphics[width=0.24\linewidth]{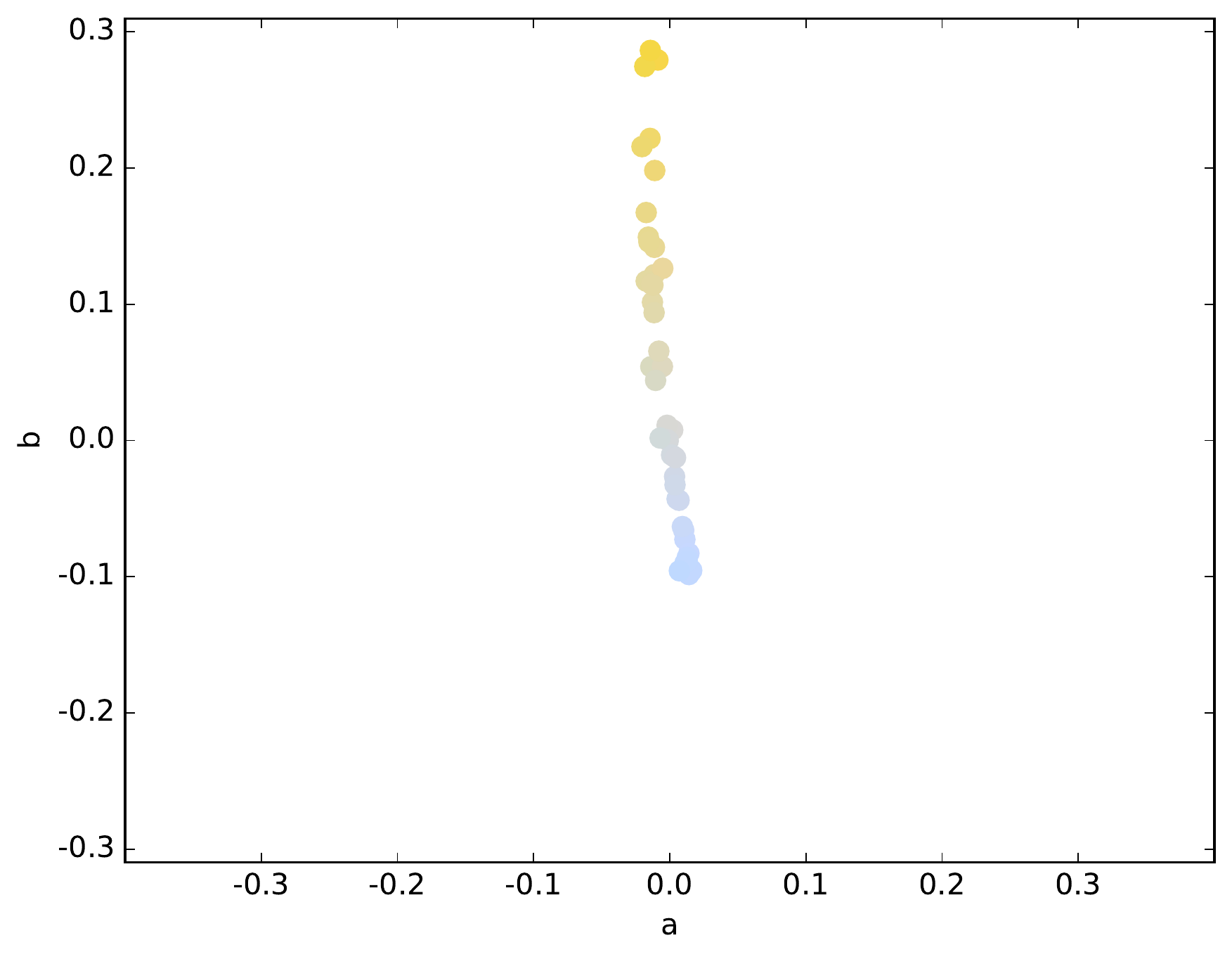}
\includegraphics[width=0.24\linewidth]{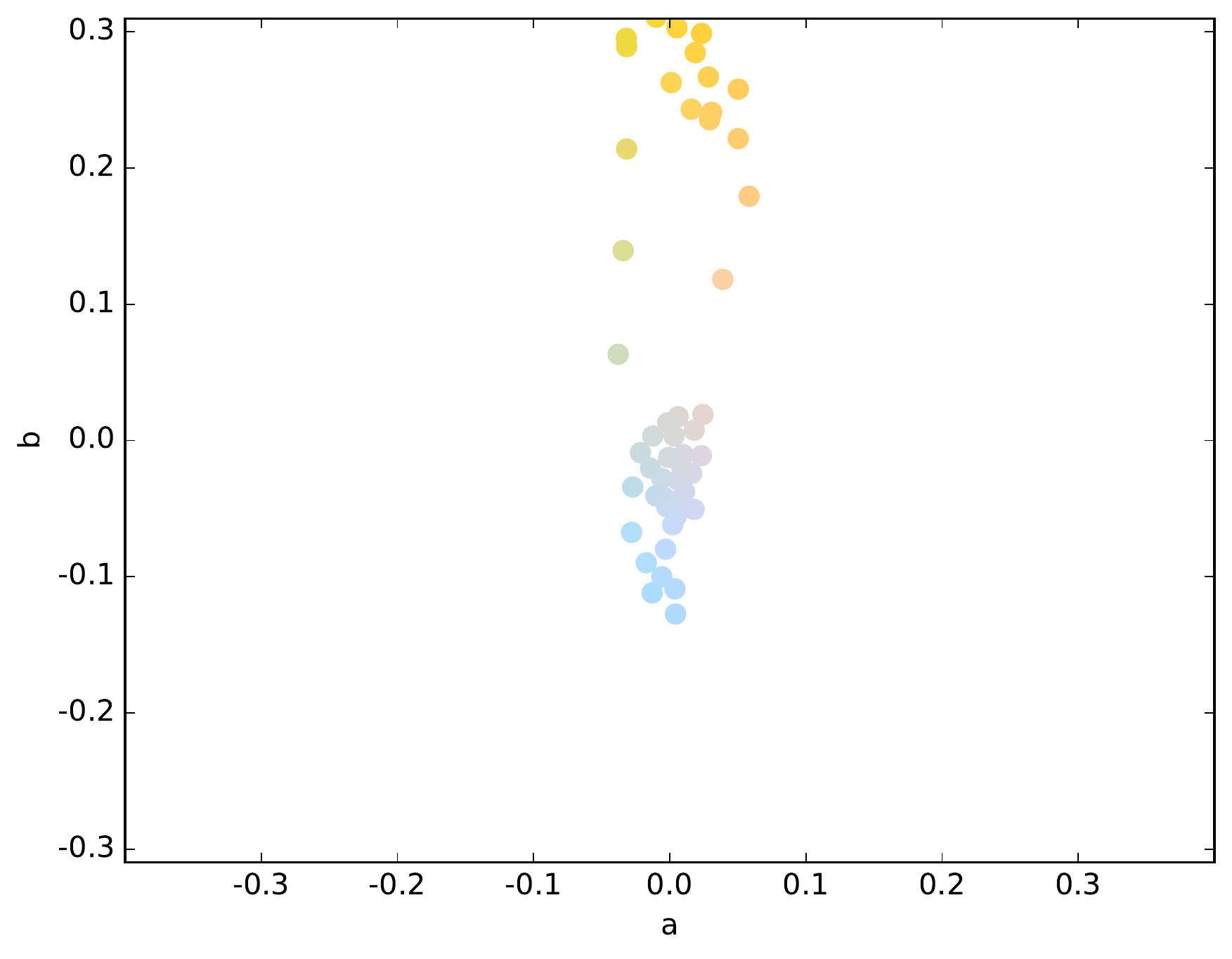}
\includegraphics[width=0.24\linewidth]{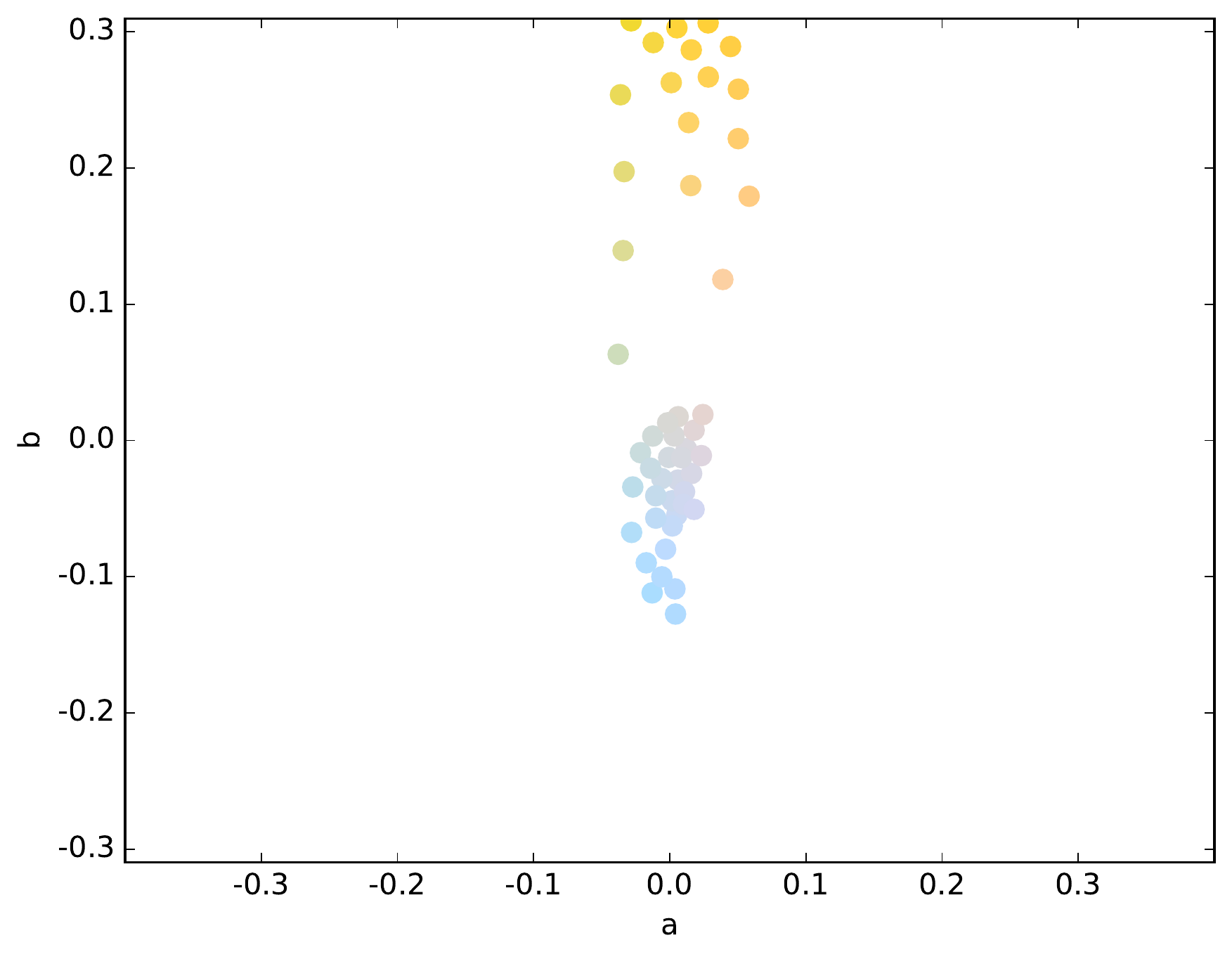}\\
\caption{\label{fig:color_transfer_palette}Examples of color transfer results where the photograph  on the left is modified to take the color of the photograph on the right. The second image from the left shows the result obtained with our graph CNN method while the second from the right shows the result obtained with optimal transport. We present below each image the corresponding color palettes $\Sig_t$, $\Sig^*$, $\Sig_{ot}$, $\Sig_s$ (from left to right).}
\end{figure*}

Beyond the transformed images, it is also interesting to study how each palette is transformed with the different methods. We present in Fig.~\ref{fig:color_transfer_palette} these different color palettes. We remark that the palette $\Sig_{ot}$ obtained with optimal transport is almost identical to the target palette $\Sig_{s}$, at the price of several artefacts in the resulting images. On the contrary, we observe more differences between the palette $\Sig^*$ obtained with our graph CNN method and $\Sig_{s}$: a better preservation of the internal structure of $\Sig_t$ is obtained in $\Sig^*$ thanks to the graph regularisation.

%

\section{ - Training graph CNNs for denoising}

\begin{figure}
\centering
\includegraphics[width=.40\linewidth]{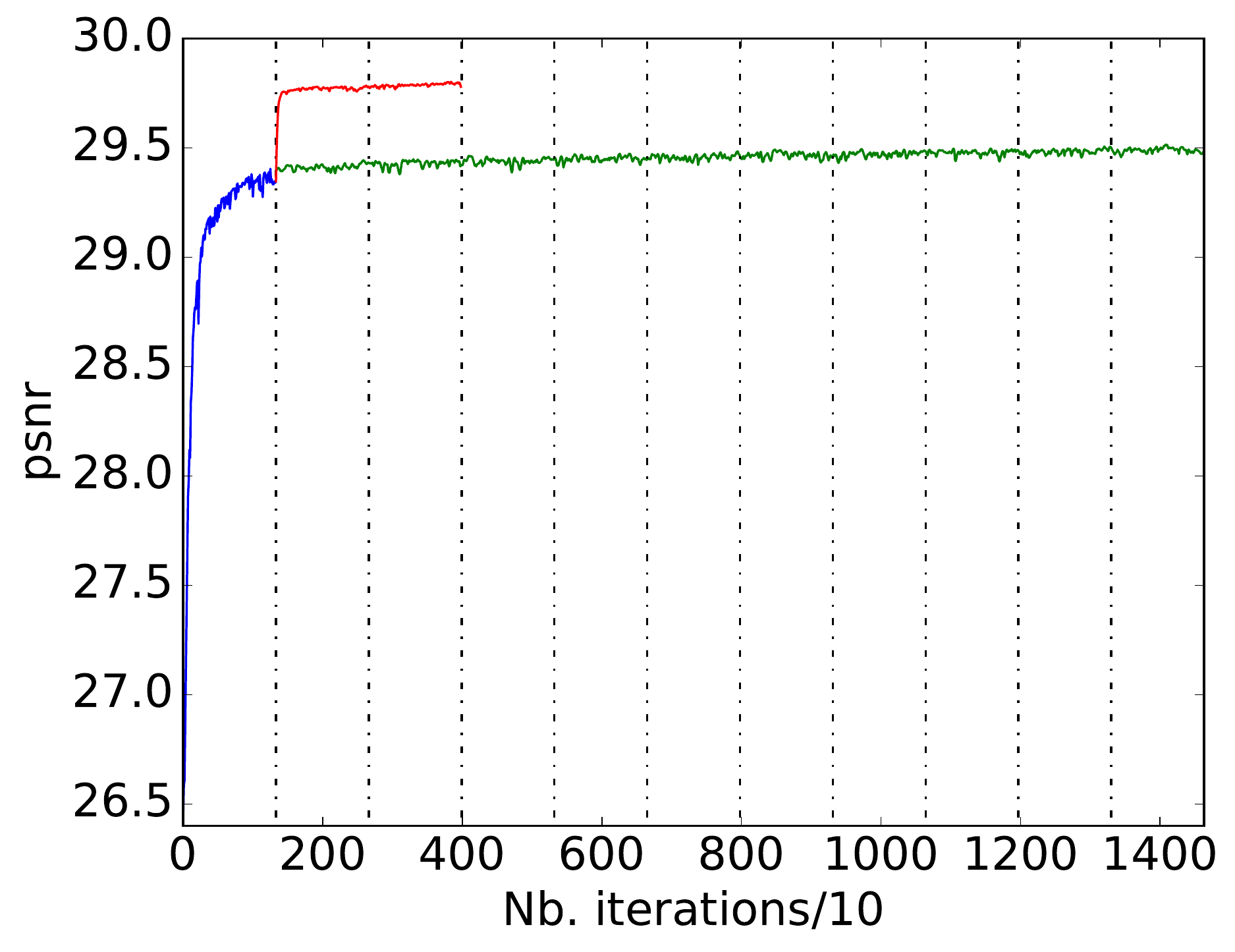}
\includegraphics[width=.40\linewidth]{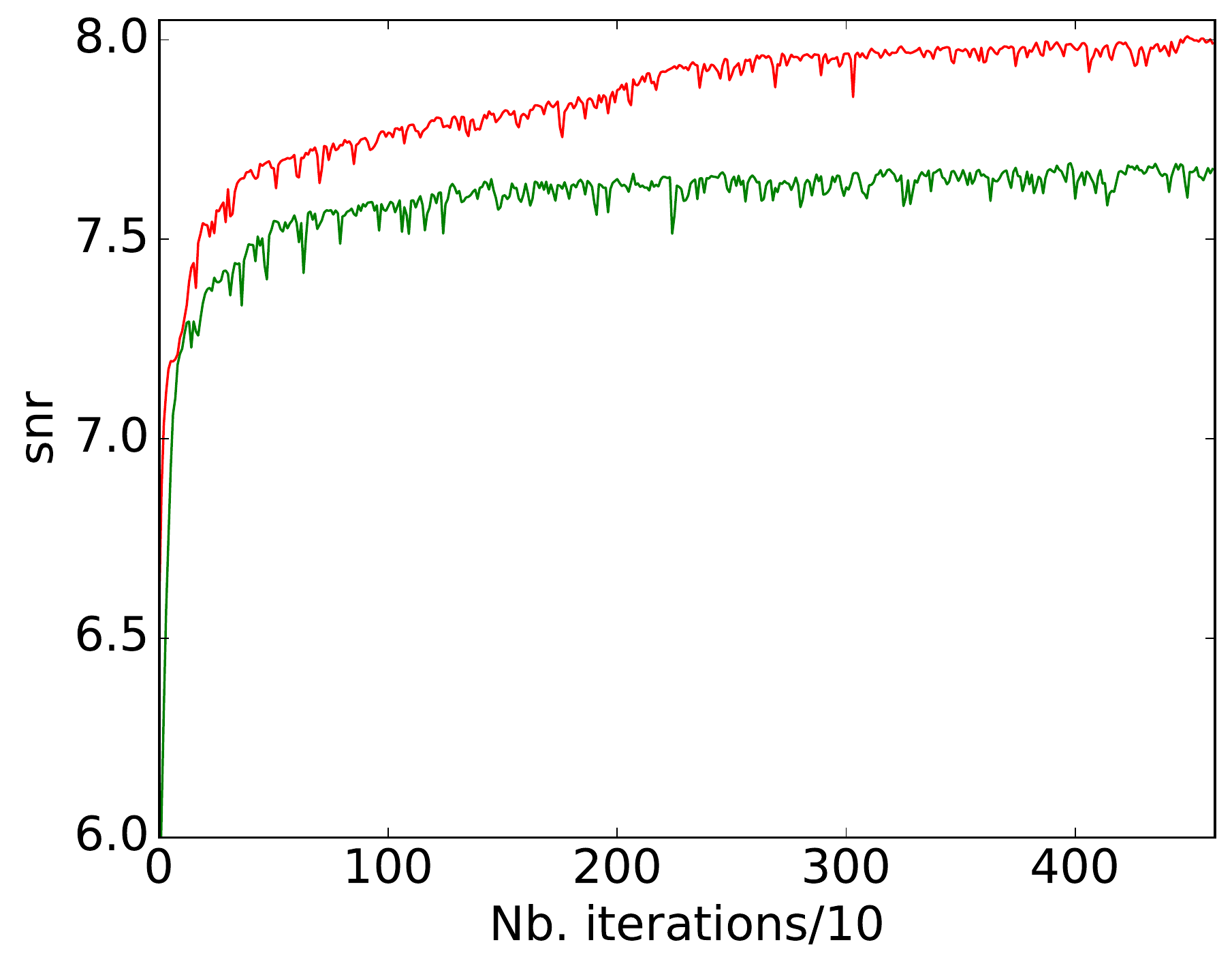}
\caption{\label{fig:evol_validation_set}Left: Evolution of the average PSNR on the image validation set during training. The blue curve corresponds to the pre-training of the local network. The green and red curves correspond to the training of the local and non-local networks, respectively, using the pre-trained network for the initialisation. The vertical dashed-dotted black lines indicate the end of an epoch. Right: Evolution of the average SNR on the audio validation set during training. The green and red curves correspond to the training of the local and non-local networks, respectively.}
\end{figure}

In this appendix, we show that one can train a graph CNN to solve standard signal processing tasks. To demonstrate that the approach can lend itself to different kinds of data, this part is about denoising images and single-channel audio signals.

The ground truth signal is denoted $\sigVec \in \Rbb^n$, and its components take values in range $[0,1]$ (image pixels), and $[-1,1]$ (audio samples). An associated noisy version satisfies
\begin{align}\label{eq:noisecorrupt}
\vec{y} = \sigVec + \vec{n},
\end{align}
where $\vec{n} \in \Rbb^\nbVert$ is a random vector drawn from the centered Gaussian distribution of standard deviation $\sigma = 0.07$, for images, or $\sigma = 0.12$, for audio. 

In the following experiments, we compare the denoising performance achieved by trained local and non-local graph CNNs.

\subsection{Network structures}

For both types of signals, the first and the last layer of denoising CNNs are using local convolutions (Section~\ref{sec:local_convolution}). We denote these layers by ${f_1 \colon \Rbb^{\nbVert} \rightarrow \Rbb^{\nbVert \times \nbFeat_1}}$ and ${f_2 \colon \Rbb^{\nbVert \times \nbFeat_1} \rightarrow \Rbb^{\nbVert}}$, where only $f_1$ incorporates a non-linearity, while $f_2$ is fully linear (without bias). For ease of notation, we slightly abuse the definition of non-linearity in \eqref{eq:layer}, to accommodate the soft-thresholding function
\begin{align}
s(x, \bias_+, \bias_-) = 
\left\{
\begin{array}{ll}
x - \bias_+ & \text{if } x \geq \bias_+\\
0 & \text{if } x \in (\bias_-, \bias_+)\\
x - \bias_- & \text{if } x \leq \bias_-
\end{array}
\right.,
\end{align}
where $\bias_+ \geq 0$ and $\bias_- \leq 0$ are parameters learned during training. The first layer $f_1$ thus takes as input a noisy signal $\vec{y}$, while $f_2$ returns its denoised version. 

The graph for the non-local convolutions is built on the noisy input signal $\vec{y}$ and thus changes with each new signal to denoise. It is constructed based on the nearest neighbour search as follows. From the noisy signal, we assemble a feature vector $\vec{f}_i \in \Rbb^{\ensuremath{r}}$ by extracting all the components (\emph{i.e.} pixels or samples) in the predefined local neighbourhood centered around the $i^\th$ entry of $\vec{y}$. We then search the $\gdeg$ nearest neighbours to $\vec{f}_i$ in the set $\{\vec{f}_1, \ldots, \vec{f}_\nbVert \}$ using the Euclidean distance. The weight $\ma{W}_{ij}$ of the adjacency matrix $\ma{W}$ between connected elements $i$ and $j$ satisfies
\begin{align}
\label{eq:weighting_NL}
\ma{W}_{ij} = \exp\bigg(- \,\alpha \; \max\bigg\{\frac{\norm{\vec{f}_i - \vec{f}_j}_2^2}{\ensuremath{r}} - \beta, \; 0 \bigg\} \bigg),
\end{align}
where $\alpha, \beta > 0$ are parameters learned during training. Note that \eqref{eq:weighting_NL} is the weighting function used in NL-means~\cite{buades11}. 

We compare the denoising performance reached by a CNN using only local convolutions, \ie, a regular CNN, and a graph CNN where we also use non-local convolutions.

\textbf{Image denoising} -- The regular CNN uses only $f_1$ and $f_2$, and, therefore, implements the function $f_2 \circ f_1$. We choose $\gdeg = 9$, hence the local filters have size $3 \times 3$, and $m_1 = 18$. The non-local network is built by inserting one non-local layer between $f_1$ and $f_2$, which we denote by $f_{\rm nl} \colon \Rbb^{\nbVert \times \nbFeat_1} \rightarrow \Rbb^{\nbVert \times \nbFeat_1}$. This network implements the function $f_{2} \circ f_{\rm nl} \circ f_{1}$. The neighbourhood used for building the graph through features is of size $7 \times 7$ ($r=49$). The non-local layer $f_{\rm nl}$ is linear (no bias or soft-thresholding) and we interpret its role as an additional non-local denoising of the sparse feature maps given by $f_{1}$ before reconstruction of the denoised image by $f_{2}$. This layer implements a 2D graph-convolution with a single filter $\vec{h} \in \Rbb^\gdeg$: $f_{\rm nl}\left(\Sig, \Graph \right)  = \left( f_1(\sigVec)_j \star \vec{h} \right)_{j = 1, \ldots, \nbFeat_1}$.
For the local layers, we use $\gdeg = 9$. We remark that, while one may expect an equivalent \emph{local} CNN (implementing $f_{2} \circ f_{\rm \ell} \circ f_{1}$, with $f_{\rm \ell}$ a fully linear 2D local convolution layer of the same size as $f_{\rm nl}$) to perform better than $f_2 \circ f_1$, our experiments showed that both yield equivalent results.

\textbf{Audio denoising} -- The audio denoising CNNs (local and non-local) share the same architecture, defined as $f_{2} \circ f_{\bullet} \circ f_{1}$. The middle layer $f_{\bullet} \colon \Rbb^{\nbVert \times \nbFeat_1} \rightarrow \Rbb^{\nbVert \times \nbFeat_1}$ is linear (no bias or soft-thresholding) and implements either local ($f_{\rm l}$) or non-local ($f_{\rm nl}$) convolutions\footnote{With slight abuse, we keep the same notation here as in the image denoising case for simplicity.} as defined in \eqref{eq:layer}. Therefore, the number of filter coefficients to train is equal in both local and non-local cases. We choose filters of size $\gdeg = 5$ and $\nbFeat_1 = 10$.
The non-local graph is generated as described before, with a local neighbourhood defined as a signal segment of length $r = 25$. 

\subsection{Experimental setup}

\begin{figure}
\includegraphics[width=.24\linewidth]{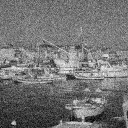}
\includegraphics[width=.24\linewidth]{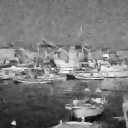}
\includegraphics[width=.24\linewidth]{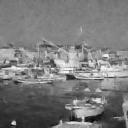}
\includegraphics[width=.24\linewidth]{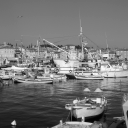}
\caption{\label{fig:example_denoising}Noisy image (left, $23.13$ dB) denoised with the trained local network (middle left, $28.28$ dB) or the trained non-local network (middle right, $28.61$ dB) and the original image (right).}
\end{figure}

In all layers, one filter is initialised to the constant filter $(1/\gdeg, \ldots, 1/\gdeg) \in \Rbb^\gdeg$, serving to extract and approximately reconstruct the mean of each patch. The remaining filters are initialised using random draws from the centered Gaussian distribution with standard deviation $0.01$. We noticed that this initialisation accelerates the learning. The  parameters $\bias_+$ and $\bias_-$ ($m_1$ of each) are initialised at $0$. 

Training is done using ADAM \cite{kingma14}, with a batchsize of size $1$, and by minimising the Euclidean distance between the denoised and ground-truth signals. 
The datasets are divided into three subsets: training, validation and test sets. The validation set is used to monitor the evolution of the PSNR (images) or SNR (audio) during training. The training and test examples are always generated by corrupting the original signal, as defined in \eqref{eq:noisecorrupt}, with independent draws of $\vec{n}$.

\textbf{Image denoising} -- We train the two image denoising CNNs using the \emph{holidays} dataset which contains $1491$ images \cite{jegou08}. The test set is built by choosing $150$ images at random among all available images; the validation set is build by choosing $15$ instances among the remaining ones; the rest form the training set. All images are cropped to become square and resized to have size $\nbVert = 128 \times 128$. First, we pretrain the local network $f_1 \circ f_2$ for $1$ epoch and a stepsize of $10^{-3}$. Second, we build the non-local network $f_{2} \circ f_{\rm nl} \circ f_{1}$ using the pre-trained layers $f_1$ and $f_2$ and train this whole network using a stepsize of $10^{-4}$. Third, starting again from the pre-trained layer $f_1$ and $f_2$, we continue training the local network $f_1 \circ f_2$ using a stepsize of $10^{-4}$. Note that training the non-local network is computationally more expensive than training the local network as a new non-local graph must be constructed for each new noisy image. We thus limited training of $f_{2} \circ f_{\rm nl} \circ f_{1}$ to $2$ epochs instead of $10$ for $f_1 \circ f_2$. For the non-local layer, the default values proposed in~\cite{buades11} are used to initialise the parameters $\alpha$ and $\beta$ ($1/(0.40\sigma^2)$ and $2\sigma^2$, respectively).

\textbf{Audio denoising} -- For audio, we use the \emph{TIMIT} \cite{garofolo1993darpa} speech dataset, counting $4618$ training and $1680$ test tracks of varying duration, sampled at $16$ kHz. The validation set is built by separating $15$ tracks from the training set chosen at random. The audio files are first trimmed to have equal length by extracting a $1$ second signal from the original audio tracks. 
The two parameters in \eqref{eq:weighting_NL} are initialised at $\alpha = 1/\sigma^2$ and $\beta=0$, and the training is performed on the whole network at once (for both local and non-local CNNs). We run the training procedure for one epoch, with the stepsize $10^{-3}$.

\subsection{Results}

\begin{figure*}
\centering
\begin{minipage}{0.49\linewidth}
\centering
Regular CNN -- $f_{2} \circ f_{1}$
\end{minipage}
\begin{minipage}{0.49\linewidth}
\centering
Graph CNN -- $f_{2} \circ f_{\rm nl} \circ f_{1}$
\end{minipage}\\
\begin{minipage}{0.24\linewidth}
\centering
Analysis layer
\end{minipage}
\begin{minipage}{0.24\linewidth}
\centering
Synthesis layer
\end{minipage}
\begin{minipage}{0.24\linewidth}
\centering
Analysis layer
\end{minipage}
\begin{minipage}{0.24\linewidth}
\centering
Synthesis layer
\end{minipage}\\
\includegraphics[width=0.24\linewidth]{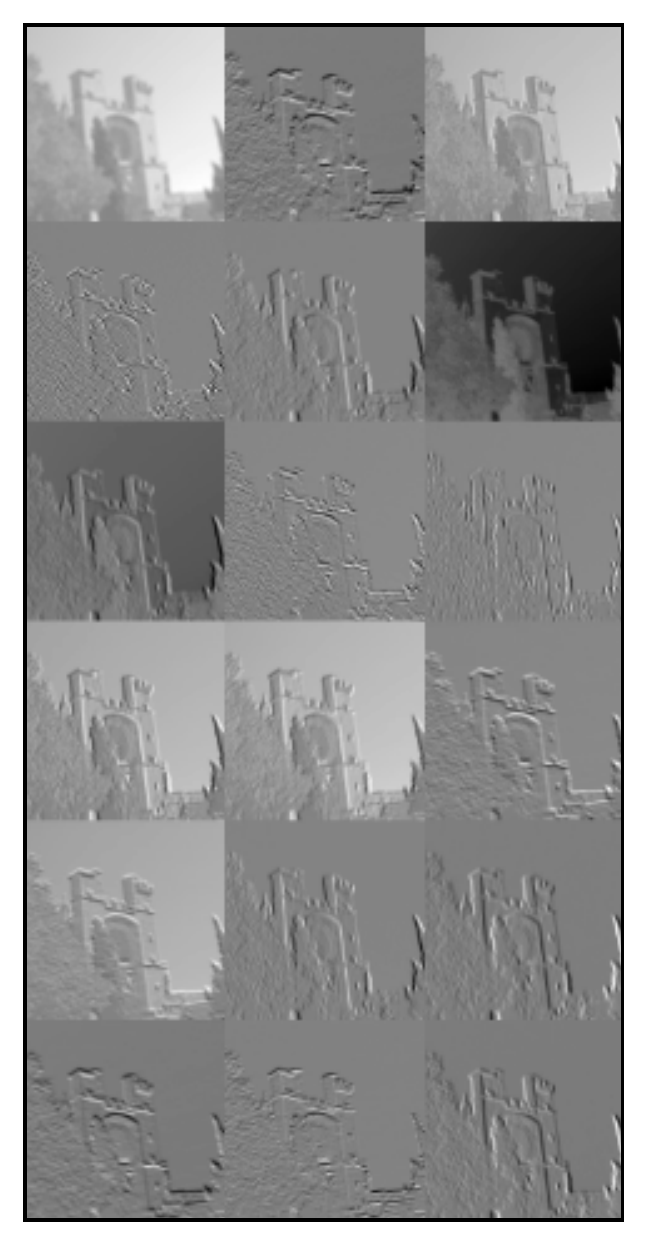}
\includegraphics[width=0.24\linewidth]{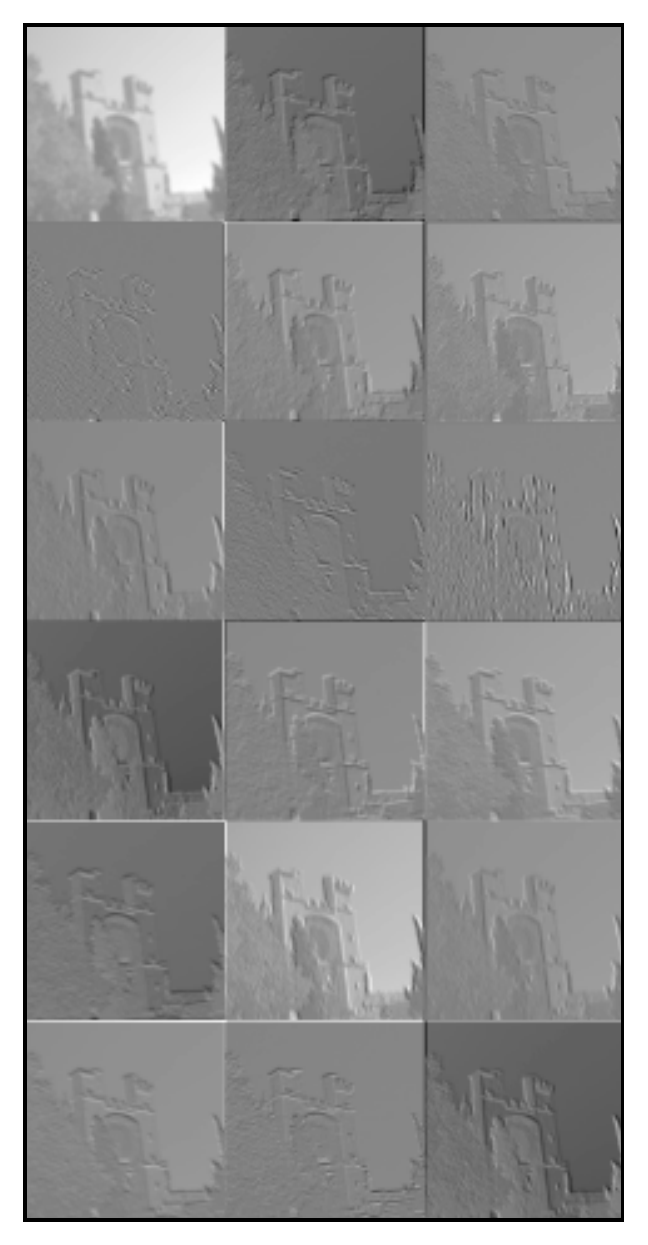}
\includegraphics[width=0.24\linewidth]{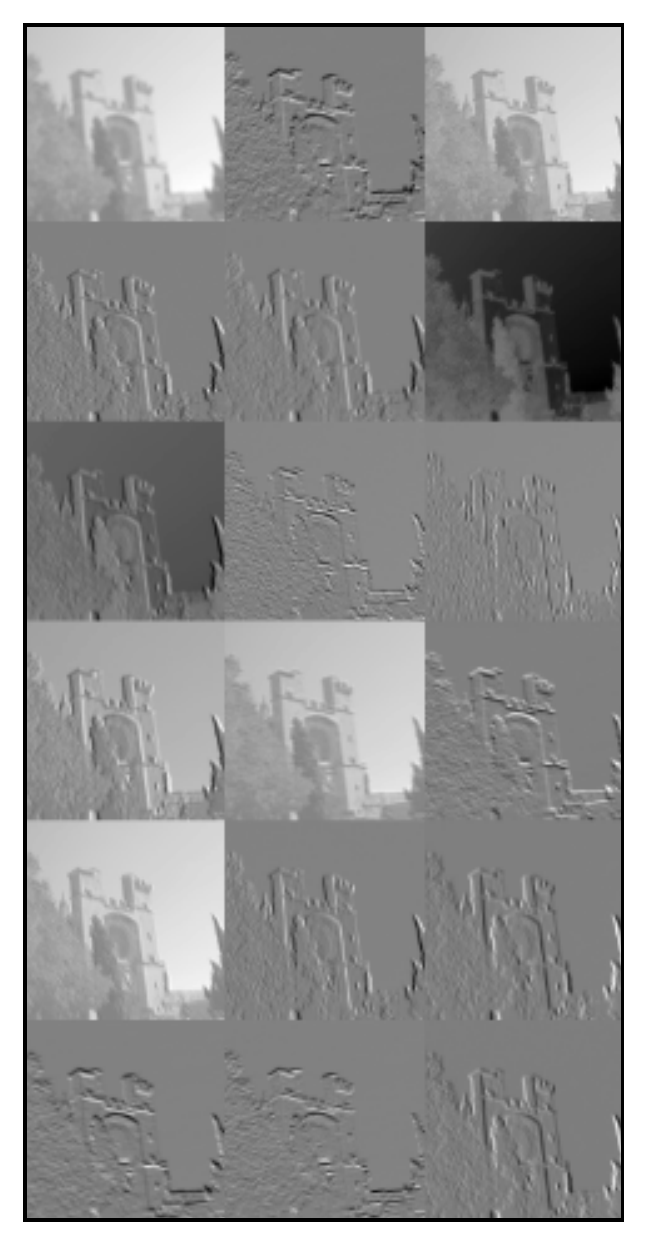}
\includegraphics[width=0.24\linewidth]{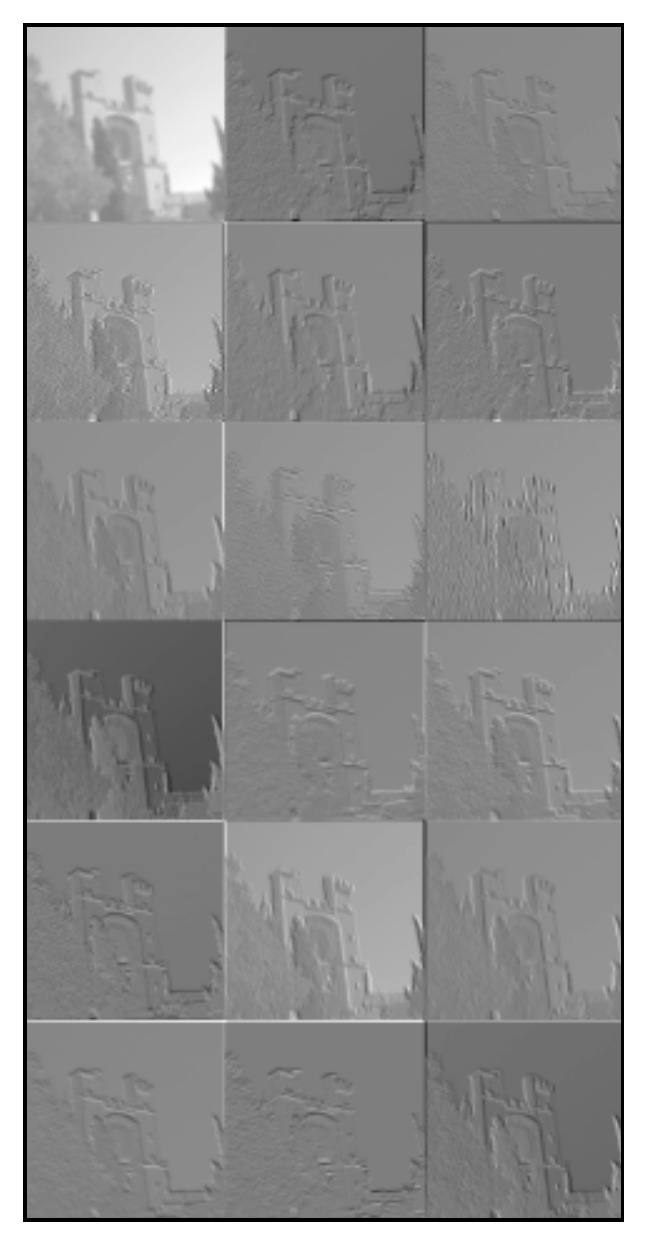}\\
\caption{\label{fig:denoising_filters}First and third panels: feature maps obtained at the analysis layer of both trained networks computed from the same image (the non-linearity was not applied). Second and fourth panels: feature maps obtained by applying the adjoint of the filters at the synthesis layer of both trained networks.}
\end{figure*}

Figure \ref{fig:evol_validation_set} presents the evolutions of the mean PSNR (for images) and the mean SNR (for audio), on the validation sets during training. One can already notice at this stage that the non-local networks outperform the local ones. 

\textbf{Image denoising} -- The average PSNR of the image test set is $23.10$ dB before denoising. After denoising, we reach a PSNR of \mbox{$29.13$ dB} and \mbox{$29.42$ dB} with the local and non-local networks, respectively. The non-local network thus improves the denoising performance by \mbox{$0.29$ dB}. For comparison, we also denoise the test set by soft-thresholding in the Haar wavelet basis. We first compute the threshold that maximises the PSNR on the validation set and then use it to denoise the test set. We reach an average PSNR of $26.78$~dB only on the image test set with this method. We present in Fig.~\ref{fig:example_denoising} an image before and after denoising with both networks. We notice that the non-local network allows a better recovery of the homogeneous regions in the image. 

By analogy with dictionary learning and the terminology used in this field, we call hereafter the first layer $f_{1}$ of the CNNs used for denoising, the ``analysis layer'', and the last layer $f_{2}$, the ``synthesis layer''. The analysis layer filters the input image with a bank of filters and soft-thresholds the result, yielding sparse feature maps. The synthesis layer resynthesises an image from the feature maps using another bank of filters. We present in Fig.~\ref{fig:denoising_filters} some feature maps obtained at the analysis layer -- without application of the non-linearity -- for both trained networks. We remark that the learned filters act like wavelet filters in both cases. This is a well-known phenomenon observed in dictionary learning or at the first layer of deep CNNs. Our results are thus in line with these observations. In addition, we remark that the introduction of the non-local convolutions did not affect a lot the type of filters learned at the first layer. We also present the feature maps obtained by applying the adjoint of the filters at the synthesis layer of both trained networks. Note that in dictionary learning, the filters at the analysis are exactly the adjoint of the filters at the synthesis. Similarly, here, we notice that the (adjoint of the) synthesis filters act like wavelet filters, just as the analysis filters do. We also remark that the synthesis filters are quite similar in absence and presence of the non-local filtering.

\textbf{Audio denoising} -- The average SNR of the audio test set is \mbox{$0.85$ dB} before denoising. Denoising by the local CNN increases the average SNR to \mbox{$8.44$ dB}, whereas, denoising with the non-local CNN increases the SNR to \mbox{$8.79$ dB}, \emph{i.e.} a \mbox{$0.35$ dB} improvement compared to the former. As for images, we also denoise the test set by soft-thresholding in the Haar wavelet basis with a threshold optimised on the validation set. We reach an average PSNR of $5.80$~dB only on the test set with this method.

\end{document}